\documentclass{article} % For LaTeX2e
\usepackage{nips13submit_e,times}
\usepackage{natbib}
\usepackage{graphicx,subfigure}
\usepackage{amsmath}
\usepackage{fancyvrb}
\usepackage{amssymb}
\usepackage{wrapfig}
\usepackage{multirow}
\usepackage{hyperref}
\usepackage{algorithm}
\usepackage{algorithmic}

\nipsfinalcopy % Uncomment for camera-ready version

\newcommand{\hess}{\mathbf{H}}
\newcommand{\jacob}{\mathbf{J}}
\newcommand{\params}{\mathbf{\theta}}
\newcommand{\cost}{\mathcal{L}}
\newcommand{\lout}{\mathbf{r}}
\newcommand{\louti}{r}
\newcommand{\outi}{y}
\newcommand{\out}{\mathbf{y}}
\newcommand{\gauss}{\mathbf{G_N}}

\newcommand{\sigmoid}{\sigma}
\newcommand{\softmax}{\phi}

\newcommand{\diag}{\text{diag}}
\newcommand{\metric}{\mathbf{F}}
\newcommand{\sample}{\mathbf{z}}

\begin{document}

\title{Revisiting natural gradient for deep networks}
\author{Razvan Pascanu\\
Universit\'e de Montr\'eal\\
Montr\'eal QC H3C 3J7 Canada\\
\texttt{r.pascanu@gmail.com} \\
\And
Yoshua Bengio\\
Universit\'e de Montr\'eal\\
Montr\'eal QC H3C 3J7 Canada\\
\texttt{yoshua.bengio@umontreal.ca}\\
}

\maketitle
\begin{abstract}
    We evaluate natural gradient descent, an algorithm originally proposed in
    \citet{Amari97}, for learning deep models.  The contributions of this paper
    are as follows.  We show the connection between natural gradient descent
    and three other recently proposed methods for training deep models:
    Hessian-Free Optimization~\citep{Martens10}, Krylov Subspace
    Descent~\citep{VinyalsP12} and TONGA~\citep{LeRoux+al-tonga-2008-small}.
    We describe how one can use unlabeled data to improve the generalization
    error obtained by natural gradient and empirically evaluate the robustness
    of the algorithm to the ordering of the training set compared to stochastic
    gradient descent.  Finally we extend natural gradient descent to
    incorporate second order information alongside the manifold information and
    provide a benchmark of the new algorithm using a truncated Newton approach
    for inverting the metric matrix instead of using a diagonal approximation
    of it. 
\end{abstract}

\vspace*{-2mm}
\section{Introduction}
\vspace*{-2mm}

Several recent papers tried to address the issue of using better optimization
techniques for machine learning, especially for training deep architectures or
neural networks of various kinds.  Hessian-Free
Optimization~\citep{Martens10,Martens11,chapelle11}, Krylov Subspace Descent
\citep{VinyalsP12, Mizutani}, natural gradient descent \citep{Amari97,Park00},
TONGA \citep{LeRoux+al-tonga-2008-small,RouxF10} are just a few of such
recently proposed algorithms. They usually can be split in different
categories: those which make use of second order information, those which use
the geometry of the underlying parameter manifold (e.g. natural gradient
descent) or those that use the  uncertainty in the gradient (e.g. TONGA). 

One particularly interesting pipeline to scale up such algorithms was
originally proposed in~\citet{Pearlmutter94} -- finetuned in
\citet{Schraudolph02} -- and represents the backbone behind both Hessian-Free
Optimization~\citep{Martens10} and Krylov Subspace Descent~\citep{VinyalsP12,
Mizutani}.  The core idea behind it is to make use of the forward pass (renamed
to {\it{R-operator}} in \citet{Pearlmutter94}) and backward pass of automatic
differentiation to compute efficient products between Jacobian or Hessian
matrices and vectors. These products are used within a  truncated-Newton
approach \citep{nocedal99} which considers the exact Hessian and only inverts
it approximately without the need for explicitly storing the matrix in memory,
as opposed to other approaches which perform a more crude approximation of the
Hessian (or Fisher) matrix (either diagonal or block-diagonal).

The original contributions of this paper to the study of natural gradient
descent are as follows.  Section~\ref{sec:hf} describes the connection between
natural gradient and Hessian Free~\citep{Martens10}, section~\ref{sec:ksd}
looks at the relationship with Krylov Subspace Descent
(KSD)~\citep{VinyalsP12}.  Section~\ref{sec:unlabeled} describes how unlabeled
data can be incorporated into natural gradient descent in order to improve
generalization error.  Section~\ref{sec:robustness} explores empirically
natural gradient descent's robustness to permutation of the training set.  In
section~\ref{sec:nncg} we extend natural gradient descent to incorporate second
order information.  Finally in section~\ref{sec:benchmark} we provide a
benchmark of the algorithm discussed, where natural gradient descent is
implemented using a truncated Newton approach for inverting the full metric
matrix instead of the traditional diagonal or band-diagonal approximation.

\vspace*{-2mm}
\section{Natural gradient descent}
\vspace*{-2mm}
\label{sec:natgrad}
Natural gradient descent can be traced back to Amari's work on information
geometry~\citep{Amari-1985} and its application to various neural
networks~\citep{Amari-1992,Amari97}, though a more in depth introduction can be
found in ~\citet{Amari98,Park00,Arnold11}. The algorithm has also been
successfully applied  in the reinforcement learning community~\citep{Kakade01,
Peters_N_2008} and for stochastic search~\citep{YiWSS09}.

Let us consider a family of density functions \mbox{$\mathcal{F}$} that maps a
parameter $\theta \in \mathbb{R}^P$ to a probability density function
$p(\sample)$, $p: \mathbb{R}^N \to [0,\infty)$, where $\sample \in
\mathbb{R}^N$.  Any choice of $\theta \in \mathbb{R}^P$ defines a particular
density function $p_{\theta}(\sample) = \mathcal{F}(\theta)(\sample)$ and by
considering all possible $\theta$ values, we explore the set $\mathcal{F}$,
which is our functional manifold.

In its infinitesimal form, the KL-divergence behaves like a distance measure,
so we can define a similarity measure between nearby density functions.  Hence
$\mathcal{F}$ is a Riemannian manifold whose metric is given by the Fisher
Information matrix $\mathbf{F}_\theta$ defined in the equation below:

\begin{equation}
\label{eq:FIM}
\mathbf{F}_\theta = \mathbb{E}_{\sample} \left[
    \left(
    \nabla \log p_{\theta}(\sample)
    \right)^T 
    \left(
    \nabla \log p_{\theta}(\sample)
    \right)
\right].
\end{equation}

That is, locally around some point $\theta$, the metric defines an inner
product between vectors $\mathbf{u}$ and $\mathbf{v}$: 
$$<\mathbf{u}, \mathbf{v}>_\theta = \mathbf{u} \mathbf{F}_\theta \mathbf{v},$$
and it hence provides a local measure of distance. It what follows we will
write $\mathbf{F}$ for the Fisher Information Matrix, assuming an implicit
dependency of this matrix on $\theta$. 

Given a loss function $\mathcal{L}$ parametrized by $\theta$, natural gradient descent
attempts to move along the manifold by correcting the gradient of $\mathcal{L}$
according to the local curvature of the KL-divergence surface, i.e. moving some given 
distance in the direction $\nabla_N \mathcal{L}(\theta)$:
\begin{align}
\label{eq:nat_grad}
\nabla_N \mathcal{L}(\theta) \buildrel \text{d{}ef}\over= &
\nabla \mathcal{L}(\theta)
\mathbb{E}_{\sample} \left[ 
    \left(
    \nabla \log p_{\theta}(\sample)
    \right)^T 
    \left(
    \nabla \log p_{\theta}(\sample)
    \right)
\right]^{-1} \nonumber \\ &
\buildrel \text{d{}ef}\over= 
\nabla \mathcal{L}(\theta) \metric^{-1}.
\end{align}

We use $\nabla_N$ for natural gradient, $\nabla$ for gradients and $\mathbf{F}$
is the metric matrix given by the Fisher Information Matrix.  Partial
derivatives are usually denoted as row vectors in this work.  We can derive
this result by considering natural gradient descent to be defined as the
algorithm which, at each step, attempts to pick a descent direction such that
the amount of change (in the KL-sense) induced in our model is some given
value.  Specifically we look for a small $\Delta \theta$ that minimizes a first
order Taylor expansion of $\mathcal{L}$ when the second order Taylor series of
the KL-divergence between $p_{\theta}$ and $p_{\theta + \Delta \theta}$ has to
be constant: 
\begin{equation}
\label{eq:nat_grad_obj}
\begin{array}{l}
\arg\min_{\Delta \theta} \mathcal{L}(\theta + \Delta \theta) \\
\text{s. t. } KL(p_{\theta}||p_{\theta + \Delta \theta}) = const.
\end{array}
\end{equation}

Using this constraint we ensure that we move along the functional manifold with
constant speed, without being slowed down by its curvature.  This also makes
learning locally \emph{robust to re-parametrizations of the model}, as the
functional behaviour of $p$ does not depend on how it is parametrized.

Assuming $\Delta \theta \to 0$, we can approximate the KL divergence by its second order Taylor series:
\begin{align}
\label{eq:KL_derived} 
KL& ( p_{\theta}  \parallel p_{\theta+\Delta \theta})  \approx 
    \left(\mathbb{E}_{\sample}\left[\log p_{\theta}\right]  
     -   \mathbb{E}_{\sample}\left[\log p_{\theta}\right] \right) \nonumber \\ 
     & -  \mathbb{E}_{\sample}\left[
    \nabla \log p_{\theta}(\sample) \right] \Delta \theta 
     - 
    \frac{1}{2}\Delta \theta^T \mathbb{E}_{\sample}\left[\nabla^2 \log p_{\theta}\right]
    \Delta \theta \nonumber \\
    & =  \frac{1}{2}\Delta\theta^T
    \mathbb{E}_{\sample}
        \left [ 
            -\nabla^2 \log p_{\theta}(\mathbf{z})
        \right]
        \Delta\theta \nonumber \\
      & =  \frac{1}{2}\Delta\theta^T
\mathbf{F}
\Delta \theta
\end{align}

The first term cancels out and because 
$\mathbb{E}_{\mathbf{z}}\left[
\nabla \log p_{\theta}(\sample)\right] = 0$\footnote{Proof:
$\mathbb{E}_{\mathbf{z}}\left[\nabla \log p_{\theta}(\sample)\right] = 
\sum_{\sample} \left(p_{\theta}(\sample) \frac{1}{p_{\theta}(\sample)}
\frac{\partial p_{\theta}(\sample)}{\partial \theta}\right) =
\frac{\partial}{\partial \theta} \left(\sum_{\theta} p_{\theta}(\sample)\right) =
\frac{\partial 1}{\partial \theta} =0$. The proof 
holds for the continuous case as well, replacing sums for integrals.}, we are
left with only the last term.  The Fisher Information Matrix form can be
obtained from the expected value of the Hessian through algebraic manipulations
(see the appendix). 

We now express equation \eqref{eq:nat_grad_obj} as a Lagrangian, where the KL
divergence is approximated by \eqref{eq:KL_derived} and $\mathcal{L}(\theta +
\Delta \theta)$ by its first order Taylor series $\mathcal{L}(\theta) + \nabla
\mathcal{L}(\theta) \Delta \theta$: 
\begin{equation}
\label{eq:lagrange}
\mathcal{L}(\theta) + \nabla \mathcal{L}(\theta)\Delta \theta + 
\frac{1}{2}\lambda \Delta \theta^T 
\mathbf{F}
  \Delta \theta
\end{equation}
Solving equation \eqref{eq:lagrange} for $\Delta \theta$ gives us the natural
gradient decent formula \eqref{eq:nat_grad}. Note that we get a scalar factor
of $2\frac{1}{\lambda}$ times the natural gradient. We fold this scalar into
the learning rate, which now also controls the weight we put on preserving the
KL-distance between $p_{\theta}$ and $p_{\theta + \Delta \theta}$.  The
approximations we use are meaningful only around $\theta$: in
\citet{Schaul2012proof} it is shown that taking large steps might harm
convergence. We deal with such issues both by using damping (i.e. setting a
trust region around $\theta$) and by properly selecting a learning rate.

\vspace*{-2mm}
\subsection{Adapting natural gradient descent for neural networks}
\vspace*{-2mm}
\label{sec:nat_grad}
In order to use natural gradient descent for deterministic neural networks we
rely on their probabilistic interpretation (see \citet{Bishop}, chapter 3.1.1
for linear regression and 4.2 for classification). For example, the output of
an MLP with linear activation function can be interpreted as the mean of a
conditional Gaussian distribution with a fixed variance, where we condition on
the input. Minimizing the squared error, under this assumption, is equivalent
to maximum likelihood. 

For classification, depending on the activation function, we can define the
output units as being success probability of each trial or event probability of
a binomial or multinoulli distribution conditioned on the input $\mathbf{x}$. 

By an abuse of notation\footnote{E.g., for softmax output layer the random
variable sampled from the multinoulli is a scalar not a vector}, we will use
$p_{\theta}(\mathbf{t}|\mathbf{x})$  to define this conditional probability
density function described above.  Because it is a conditional probability
function, the formulation of natural gradient descent, equation
\eqref{eq:nat_grad_obj}, changes into the following equation:
\begin{equation}
\begin{array}{l}
\ \arg\min_{\Delta \theta} \mathcal{L}(\theta + \Delta \theta) \\
\text{s. t. } \mathbb{E}_{\mathbf{x} \sim \tilde{q}(\mathbf{x})}\left[KL(p_{\theta}(\mathbf{t}|\mathbf{x})||p_{\theta + \Delta \theta}(\mathbf{t}|\mathbf{x}))\right] = const.
\end{array}
\label{eq:cond_nat_grad_obj}
\end{equation}

Each value of $\mathbf{x}$ now defines a different family of density functions
$p_{\theta}(\mathbf{t}|\mathbf{x})$, and hence a different manifold.  In order
to measure the functional behaviour of $p_{\theta}(\mathbf{t}|\mathbf{x})$ for
different values of $\mathbf{x}$, we use the expected value (over $x$) of the
KL-divergence between $p_{\theta}(\mathbf{t}|\mathbf{x})$ and $p_{\theta+\Delta
\theta}(\mathbf{t}|\mathbf{x})$.

In defining the constraint of equation \ref{eq:cond_nat_grad_obj}, we have chosen to
allow ourselves the freedom to compute the expectation over $\mathbf{x}$ using
some distribution $\tilde{q}$ instead of the empirical distribution $q$.
Usually we want $\tilde{q}$ to be $q$, though one can imagine situations when
this would not be true. E.g.  when we want our model to look more carefully
at certain types of inputs, which we can do by biasing $\tilde{q}$ towards that
type of inputs.

Applying the same steps as in section \ref{sec:natgrad} we can recover the
formula for natural gradient descent.  This formula can be massaged further (similar to
\cite{Park00}) for specific activations and error functions.  We exclude these
derivations due to space constraints (they are provided in the appendix).
For convenience we show the formulas for two typical pairs of output
activations and corresponding error function, where $\mathbf{J}_{\mathbf{y}}$
stands for the Jacobian matrix $\frac{\partial \mathbf{y}}{\partial \theta}$,
$\mathbf{y}$ being the output layer. By an abuse of notation, $\frac{1}{\mathbf{y}}$ 
indicates the element-wise division of $1$ by $\mathbf{y}_i$. 
\begin{align}
\label{eq:deriv_FIM_linear}
\metric_{linear} & = \beta^2 \mathbb{E}_{\mathbf{x} \sim \tilde{q}} \left[ \frac{\partial \out}{\partial \theta}^T \frac{\partial \out}{\partial \theta} \right] 
  = \beta^2 \mathbb{E}_{\mathbf{x} \sim \tilde{q}} \left[ \jacob_{\out}^T \jacob_{\out} \right] \\
\label{eq:sigmoid_FIM}
\metric_{sigmoid} & = \mathbb{E}_{\mathbf{x} \sim \tilde{q}} \left[
\jacob_{\out}^T \diag\left(\frac{1}{\out (1-\out)}\right) \jacob_{\out} \right] 
\end{align}

\vspace*{-2mm}
\section{Properties of natural gradient descent}
\vspace*{-2mm}
\label{sec:prop}

In this section we summarize a few properties of natural gradient descent that
have been discussed in the literature. 

\emph{Natural gradient descent can be used in the online regime}.  We assume
that even though we only have a single example available at each time step one
also has access to a sufficiently large set of held out examples. If we are to
apply a second order method in this regime,  computing the gradient on the
single example and the Hessian on the held out set would be conceptually
problematic.  The gradient and Hessian will be incompatible as they do not
refer to the same function.  However, for natural gradient descent, the metric
comes from evaluating an independent expectation that is not related to the
prediction error. It measures an intrinsic property of the model. It is
therefore easier to motivate using a held-out set (which can even be unlabeled
data as discussed in section \ref{sec:unlabeled}).

\citet{DesjardinsPascanu13} shows a straight forward application of natural
gradient to Deep Boltzmann Machines.  It is not obvious how the same can be
done for a standard second order method.  Probabilistic models like RBMs and
DBMs not have direct access to the cost they are minimizing, something natural
gradient does not require.

\emph{Natural gradient descent is robust to local re-parametrization of the
model}.  This comes from the constraint that we use. The KL-divergence is a
measure of how the probability density function changes, regardless on how it
was parametrized. \citet{Dickstein12} explores this idea, defining natural
gradient descent as doing whitening in the parameter space.

The metric $\mathbf{F}$  has two different forms as can be seen in equation
\eqref{eq:KL_derived}.  Note however that while the metric can be seen as both
a Hessian or as a covariance matrix, \emph{it is not the Hessian of the cost,
nor the covariance of the gradients we follow to a local minimum}. The
gradients are of $p_\theta$ which acts as a proxy for the cost $\mathcal{L}$.
The KL-surface considered at any point $\mathbf{p}$ during training always has
its minimum at $\mathbf{p}_{\theta}$, and the metric we obtain is always
positive semi-definite by construction which is not true for the Hessian of
$\mathcal{L}$.

Because $\mathbf{F}$ can be interpreted as an expected Hessian, it measures how
much a change in $\theta$ will affect the gradients of
$\mathbb{E}_{\mathbf{t}\sim p(\mathbf{t}|\mathbf{x})}\left[\log
p_{\theta}(\mathbf{t}|\mathbf{x})\right]$.  This means that, in the same way
second order methods do, \emph{natural gradient descent can jump over plateaus}
of $p_{\theta}(\mathbf{t}|\mathbf{x})$. Given the usual form of the loss
function $\mathcal{L}$, which is just an evaluation of $p$ for certain pairs
$(\mathbf{x},\mathbf{t})$, plateaus of $p$ usually match those of $\mathcal{L}$
and hence the method can jump over plateaus of the error function.  

If we look at the KL constraint that we enforce at each time step, it does not
only ensure that we induce at least some epsilon change in the model, but also
that the model does not change by more than epsilon.  We argue that this
provides some kind of robustness to overfitting. The model is not allowed to
move \emph{too far} in some direction $\mathbf{d}$ if moving along $\mathbf{d}$
changes the density computed by the model substantially.

\vspace*{-2mm}
\section{Natural gradient descent and TONGA}
\vspace*{-2mm}
\label{sec:tonga}
In \citet{LeRoux+al-tonga-2008-small} one assumes that the gradients computed
over different minibatches are distributed according to a Gaussian centered
around the true gradient with some covariance matrix $\mathbf{C}$.  By using
the uncertainty provided by $\mathbf{C}$ we can correct the step that we are
taking to maximize the probability of a downward move in generalization error
(expected negative log-likelihood), resulting in a formula similar to that of
natural gradient descent. 

While the probabilistic derivation requires the centered covariance, in
\citet{LeRoux+al-tonga-2008-small} it is argued that one can use the uncentered
covariance $\mathbf{U}$ resulting in a simplified formula which is sometimes
confused with the metric derived by Amari:
\begin{equation}
\label{eq:uncentered_leroux}
\begin{array}{lll}
\mathbf{U} 
\approx
\mathbb{E}_{(\mathbf{x}, \mathbf{t}) \sim q} \left[
\left(
\frac{\partial \log p(\mathbf{t}|\mathbf{x})}{\partial \theta}
 \right)^T
\left(\frac{ \partial \log p(\mathbf{t}|\mathbf{x})}{\partial \theta}
\right)\right] \\
\end{array}
\end{equation}

The discrepancy comes from the fact that the equation is an expectation, though
the expectation is \emph{over the empirical distribution $q(\mathbf{x},
\mathbf{t})$ as opposed to $\mathbf{x}\sim q(\mathbf{x})$ and $ \mathbf{t}\sim
p(\mathbf{t}|\mathbf{x})$}.  It is therefore not clear if $\mathbf{U}$ tells us
how $p_\theta$ would change, whereas it is clear that  Amari's metric does.

\vspace*{-2mm}
\section{Natural gradient descent and Hessian-Free Optimization}
\label{sec:hf}
\vspace*{-2mm}

Hessian-Free Optimization as well as  Krylov Subspace Descent\footnote{By an
abuse of languge, the term Hessian-Free Optimization refers to the specific
algorithm proposed by \citet{Martens10}, whily KSD stands for the specific
algorithm proposed by \cite{VinyalsP12}} rely on the \emph{extended
Gauss-Newton approximation of the Hessian}, $\gauss$, instead of the actual
Hessian (see  \cite{Schraudolph02}):
\begin{align}
    \gauss &= \frac{1}{n} \sum_{i=1}^n \left[
 \left(\frac{\partial \lout}{\partial \params}\right)^T 
         \frac{\partial^2 \log p(\mathbf{t}^{(i)}|\mathbf{x}^{(i)})}{\partial \mathbf{r}^2} 
        \left(\frac{\partial \lout}{\partial \params} \right)\right] \nonumber \\
    &= \mathbb{E}_{\mathbf{x} \sim \tilde{q}} \left[
        \jacob_{\lout}^T \left(\mathbb{E}_{\mathbf{t} \sim \tilde{q}(\mathbf{t}|\mathbf{x})} \left[\hess_{\cost \circ \lout}\right]\right) \jacob_{\lout} \right]
\label{eq:gn_general_formula}
\end{align}
The last step of equation \eqref{eq:gn_general_formula} assumes that
$(\mathbf{x}^{(i)}, \mathbf{t}^{(i)})$ are i.i.d samples, and $\tilde{q}$
stands for the distribution represented by the minibatch over which the matrix
is computed.  $\jacob$ stands for Jacobian and $\hess$ for Hessian. The
subscript describes for which variable the quantity is computed. A composition
in the subscript, as in $\hess_{\cost \circ \lout}$, implies computing the
Hessian of $\cost$ with respect to $\lout$, with $\lout$ being the output layer
before applying the activation function.

The reason for choosing this approximation over the Hessian is not
computational, as computing both can be done equally fast. The extended
Gauss-Newton approximation is  better behaved during learning.  This is assumed
to hold because $\gauss$ is positive semi-definite by construction, so one
needs not worry about negative curvature issues.

It is known that the Gauss-Newton approximation (for linear activation function
and square error) matches the Fisher Information matrix. In this section we
show that \emph{this identity holds also for other matching pairs like sigmoid
    and cross-entropy or softmax and negative log-likelihood for which the
extended Gauss-Newton is defined}.  By choosing this specific approximation,
one can therefore view both \emph{Hessian-Free Optimization and KSD as being
implementations of natural gradient descent}.  We make the additional note that
\cite{hes99b} makes similar algebraic manipulations as the ones provided in
this section, however for different reasons, namely to provide a new
justification of the algorithm that relies on distance measures.  The original
contribution of this section is in describing the relationship between
Hessian-Free Optimization and Krylov Subspace Descent on one side and natural
gradient descent on the other. This relation was not acknowledged anywhere in
the literature as far as we are aware of.  While \cite{hes99b} precedes both
\cite{Schraudolph02, Martens10} both Hessian Free and Krylov Subspace Descent
are introduced as purely approximations to second order methods.

In the case of sigmoid units with cross-entropy objective ($\sigmoid$ is the
sigmoid function), $\hess_{\cost \circ \lout}$ is 
\begin{equation}
      \label{eq:deriv_sigmoid_gn}
    \begin{array}{lll}
         \hess_{{\cost \circ \lout}_{ij, i \neq j}} 
        & =  \frac{\partial^2 \sum_k \left(-t_k \log(\sigmoid(\louti_k)) - (1-t_k)\log(1 - \sigmoid(\louti_k)) \right)}{\partial \louti_i \partial \louti_j} \\
         & =  \frac{\partial \sigmoid(\louti_i) - t_i}{\partial \louti_j} %\\
          =  0 \\
        \hess_{{\cost \circ \lout}_{ii}} 
        & =  ... %\\
         =  \frac{\partial \sigmoid(\louti_i) - t_i}{\partial \louti_i} %\\
          =   \sigmoid(\louti_i) (1-\sigmoid(\louti_i))
    \end{array}
\end{equation}

If we insert this back into the Gauss-Newton approximation of the Hessian and
re-write the equation in terms of $\jacob_{\out}$ instead of $\jacob_{\lout}$,
we get the corresponding natural gradient metric, equation
\eqref{eq:sigmoid_FIM}. $diag(\mathbf{v})$ stands for the diagonal matrix
constructed from the values of the vector $\mathbf{v}$ and we make an abuse of
notation, where by $\frac{1}{\mathbf{y}}$ we understand the vector obtain by
applying the division element-wise (the $i$-th element of is
$\frac{1}{\mathbf{y}_i}$).

\begin{equation}
     \label{eq:sigmoid_gn}
    \begin{array}{lll}
        \gauss &= \frac{1}{n} \sum_{\mathbf{x}^{(i)}, \mathbf{t}^{(i)}} \jacob_{\lout}^T \hess_{\cost \circ \lout} \jacob_{\lout} \\
             &  =  \frac{1}{n} \sum_{\mathbf{x}^{(i)}} \jacob_{\lout}^T 
             diag\left(\out (1 - \out)\right) 
             diag\left(\frac{1}{\out (1 - \out)}\right) diag\left(\out (1 - \out)\right) \jacob_{\lout} \\
             &=\mathbb{E}_{\mathbf{x} \sim \tilde{q}}\left[ \jacob_{\out}^T diag\left(\frac{1}{\out (1 - \out)}\right) \jacob_{\out}\right] \\
    \end{array}
\end{equation}

The last matching activation and error function that we consider is the softmax
with cross-entropy.  The derivation of the Gauss-Newton approximation is given
in equation \eqref{eq:deriv_softmax_gn}. 

\begin{equation}
    \label{eq:deriv_softmax_gn}
    \begin{array}{lll}
        \hess_{{\cost \circ \lout}_{ij, i \neq j}} 
        & = & \frac{\partial^2 \sum_k \left(-t_k \log(\softmax(\louti_k))\right)}{\partial \louti_i \partial \louti_j}  
         =  \frac{\partial \sum_k \left(t_k \softmax(\louti_i)\right) - t_i }{\partial \louti_j} \\ 
         &=&  -\softmax(\louti_i) \softmax(\louti_j) \\
        \hess_{{\cost \circ \lout}_{ii}} 
        & = & ... %\\
         =  \frac{\partial \softmax(\louti_i) - t_i}{\partial \louti_i} %\\
         =  \softmax(\louti_i) - \softmax(\louti_i) \softmax(\louti_i) \\
    \end{array}
\end{equation}

\begin{equation}
    \label{eq:deriv_softmax_natgrad}
    \begin{array}{ll}
    \metric &= \mathbb{E}_{\mathbf{x} \sim \tilde{q}}\left[ \sum_{k=1}^o \frac{1}{y_k} \left(\frac{\partial y_k}{\partial \theta}\right)^T\frac{\partial y_k}{\partial \theta}\right] \\ 
     &= \mathbb{E}_{\mathbf{x} \sim \tilde{q}}\left[ \jacob_{\lout}^T \left( \sum_{k=1}^o \frac{1}{y_k} \left( \frac{\partial y_k}{\partial \lout}\right)^T \left(\frac{\partial y_k}{\partial \lout}\right) \right) 
    \jacob_{\lout} \right] \\
     &= \frac{1}{N} \sum_{\mathbf{x}^{(i)}} \left( \jacob_{\lout}^T \mathbf{M} \jacob_{\lout} \right)
    \end{array}
\end{equation}
\begin{equation}
    \label{eq:deriv_M}
    \begin{array}{ll}
   \mathbf{M}_{ij, i \neq j} & = \sum_{k=1}^o \frac{1}{y_k}\frac{\partial y_k}{\partial \louti_i} \frac{\partial y_k}{\partial \louti_j} 
                              =  \sum_{k=1}^o (\delta_{ki} - y_i)y_k(\delta_{kj} - y_j) \\
                              &=  y_iy_j - y_iy_j - y_i y_j 
                               =  - \softmax(\louti_i) \softmax(\louti_j) \\
  \mathbf{M}_{ii}  = & \sum_{k=1}^o \frac{1}{y_k}\frac{\partial y_k}{\partial y_i} \frac{\partial y_k}{\partial \louti_j} 
                   =  y_i^2 \left(\sum_{k=1}^o y_k \right) + y_i -2y_i^2  \\
                   & =  \softmax(\louti_i) - \softmax(\louti_i) \softmax(\louti_i) \\
    \end{array}
\end{equation}

Equation \eqref{eq:deriv_softmax_natgrad} starts from the natural gradient
metric and singles out a matrix $\mathbf{M}$ in the formula such that the
metric can be re-written as the product $\jacob_{\lout}^T \mathbf{M}
\jacob_{\lout}$ (similar to the formula for the Gauss-Newton approximation). In
equation \eqref{eq:deriv_M} we show that indeed $\mathbf{M}$ equals
$\hess_{\cost \circ \lout}$  and hence the natural gradient metric is the same
as the extended Gauss-Newton matrix for this case as well. Note that $\delta$
is the Kronecker delta, where $\delta_{ij, i \neq j} = 0$ and $\delta_{ii} =
1$.

There is also a one to one mapping for most of the other heuristics used by
Hessian-Free Optimization.  Following the functional manifold interpretation of
the algorithm, we can recover the Levenberg-Marquardt heuristic used in
\citet{Martens10}. This is a heuristic for adapting $\alpha$ in the damping
term $\alpha \mathbf{I}$ that is added to the Hessian before inverting it.
This additive term helps making the matrix easier to invert (for example when
it is close to singular) or to deal with negative curvature. $\mathbf{I}$ is
the identity matrix, $\alpha \in \mathbb{R}$ is a scalar. 

The justification of the method comes from a trust region approach. We look at
how well our second order Taylor approximation of the function predicts the
change when taking a step. If it does well, then we can trust our approximation
(and decrease the damping $\alpha$). If our approximation does not predict well
the change in the error, we increase the damping factor. We can see that in the
limit, when $\alpha \rightarrow$ 0, we completely trust our approximation and
use no damping.  When $\alpha$ is very large, there are two things happening.
First of all, we are taking much smaller steps (scaled by $\frac{1}{\alpha}$).
Secondly, the direction that we follow is mostly that of the gradient (ignoring
the curvature information of the Hessian). 

A similar argument can be carried on for natural gradient descent, where we
consider only a first order Taylor approximation on the manifold. For any function $f$, if
$\mathbf{p}$ depicts the picked descent direction and $\eta$ the step size

\begin{equation}
    f\left(\theta_t - \eta \mathbf{p}\right) \approx
    f(\theta_t) - \eta \frac{\partial f(\theta_t)}{\partial \theta_t} \mathbf{p}
\end{equation}

This gives the reduction ratio given by equation \eqref{eq:our_rho2} which can,
under the assumption that $p = \frac{\partial f(\theta_t)}{\partial \theta_t} \mathbf{F}^{-1}$,
be shown to behave identically with the one in \citet{Martens10} (under the same assumption, namely 
that CG is close to convergence).
\begin{equation}
\label{eq:our_rho2}
\rho = \frac{f\left(\theta_t - \eta \mathbf{p}\right) - f(\theta_t)}
{ - \eta \frac{\partial f(\theta_t)}{\partial \theta_t} \mathbf{p}} 
\approx
\frac{f(\left(\theta-t - \eta
\mathbf{F}^{-1} \frac{\partial f(\theta_t)}{\partial \theta_t}^T\right) -  f(\theta_t)}
{- \eta \frac{\partial f(\theta_t)}{\partial \theta_t} \mathbf{F}^{-1} \frac{\partial f(\theta_t)}{\partial \theta_t}^T}
\end{equation}
Structural damping \citep{Martens11}, a specific regularization term used to
improve training of recurrent neural network, can also be explained from the
natural gradient descent perspective.  Roughly it implies using the joint
probability density function $p(\mathbf{t},\mathbf{h}|\mathbf{x})$, where
$\mathbf{h}$ is the hidden state, when writing the KL-constraint.  $\log
p(\mathbf{t},\mathbf{h}|\mathbf{x})$ will break in the sum of two terms, one
being the Fisher Information Matrix, while the other measures the change in
$\mathbf{h}$ and forms the structural damping term. While theoretically
pleasing, however this derivation results in a fixed coefficient of 1 for the
regularization term.

We can fix this by using two constraints when deriving the natural gradient
descent algorithm, namely:

\begin{equation}
\label{eq:hf_cond_nat_grad_obj}
\begin{array}{l}
\ \arg\min_{\Delta \theta} \mathcal{L}(\theta + \Delta \theta) \\
\text{s. t. } \mathbb{E}_{\mathbf{x} \sim \tilde{q}(\mathbf{x})}\left[KL(p_{\theta}(\mathbf{t}|\mathbf{x})||p_{\theta + \Delta \theta}(\mathbf{t}|\mathbf{x}))\right] = const.\\
\text{and } \mathbb{E}_{\mathbf{x} \sim \tilde{q}(\mathbf{x})}\left[KL(p_{\theta}(\mathbf{h}|\mathbf{x})||p_{\theta + \Delta \theta}(\mathbf{h}|\mathbf{x}))\right] = const.
\end{array}
\end{equation}

If we apply the same steps as before for both constraints (i.e. replace them by
a second order Taylor expansion), the second term will give rise to the
structural damping term.

\vspace*{-2mm}
\section{Natural gradient descent and Krylov Subspace Descent}
\vspace*{-2mm}
\label{sec:ksd}
Instead of using linear conjugate gradient descent for computing the inverse of
the metric, Krylov Subspace Descent(KSD)~\cite{VinyalsP12} \footnote{KSD stands
for the exact algorithm proposed by \citet{VinyalsP12}} opts for restricting
$\Delta \theta$ to a lower dimensional Krylov subspace given by
$\mathbf{F}\mathbf{x} = \nabla \mathcal{L}$ and then, using some other second
order method like BFGS, to solve for $\Delta \theta$ within this space.

Formally we are looking for $\gamma_1, \gamma_2, .., \gamma_k$ such that:
\begin{equation}
 \min_{\mathbf{\gamma}} \mathcal{L}\left(\theta + \left[
 \begin{array}{l} \gamma_1 \\
 \gamma_2 \\ \ldots \\ \gamma_k \end{array} \right]
 \left[ \begin{array}{l}
  \nabla \mathcal{L} \\
  \nabla \mathcal{L} \mathbf{F} \\
  \ldots \\
  \nabla \mathcal{L} \mathbf{F}^{k-1} \end{array} \right] \right)
\end{equation}

Because it relies on the extended Gauss-Newton approximation of the Hessian,
like Hessian-Free Optimization, KSD implements natural gradient descent. But
there is an additional difference between KSD and HF that can be interpreted
from the natural gradient descent perspective. 

In order to mimic Hessian-Free Optimization's warm restart, this method adds to
the Krylov Subspace the previous search direction.  We hypothesize that due to
this change, KSD is more similar to natural conjugate gradient than natural
gradient. Natural conjugate gradient (see section \ref{sec:nncg}) is an
extension of the natural gradient descent algorithm that incorporates second
order information by applying the nonlinear conjugate gradient algorithm on top
of the natural gradients. 

To show this we can rewrite the new subspace as:
\begin{equation}
\min \mathcal{L}\left( \theta + \beta d_{t-1} + \alpha \left[
\begin{array}{l}
\frac{\gamma_1}{\alpha} \\
\frac{\gamma_2}{\alpha} \\
\ldots \\
\frac{\gamma_k}{\alpha} \end{array} \right]
\left[ \begin{array}{l}
\nabla \mathcal{L} \\
\nabla \mathcal{L}\mathbf{F} \\
\ldots \\
\nabla \mathcal{L} \mathbf{F}^{k-1} \end{array} \right]\right)
\end{equation}

From this formulation one can see that the previous direction plays a different
role than the one played when doing warm restart of CG. The algorithm is
reminiscent of the nonlinear conjugate gradient. The descent directions we
pick, besides incorporating the natural gradient direction, also tend to be
locally conjugate to the Hessian of the error with respect to the functional
behaviour of the model. Additionally, compared to nonlinear conjugate gradient
BFGS is used to compute $\beta$ rather then some known formula.

\vspace*{-2mm}
\section{Using unlabeled data}
\vspace*{-2mm}
\label{sec:unlabeled}

When computing the metric of natural gradient descent, the expectation over the
target $\mathbf{t}$ is computed where $\mathbf{t}$ is taken from the model
distribution for some given $\mathbf{x}$: $\mathbf{t}\sim
p(\mathbf{t}|\mathbf{x})$. For the standard neural network models this
expectation can be evaluated analytically (given the form of
$p(\mathbf{t}|\mathbf{x})$).  This means that we do not need target values to
compute the metric of natural gradient descent. 

Furthermore, to compute the natural gradient descent direction we need to
evaluate two different expectations over $\mathbf{x}$. The first one is when we
evaluate the expected (Euclidean) gradient, while the second is when we
evaluate the metric.  In this section we explore the effect of re-using the
same samples in computing these two expectations as well as the effect of
improving the accuracy of the metric $\mathbf{F}$ by employing unlabeled data. 

\begin{figure}
\includegraphics[width=.5\textwidth]{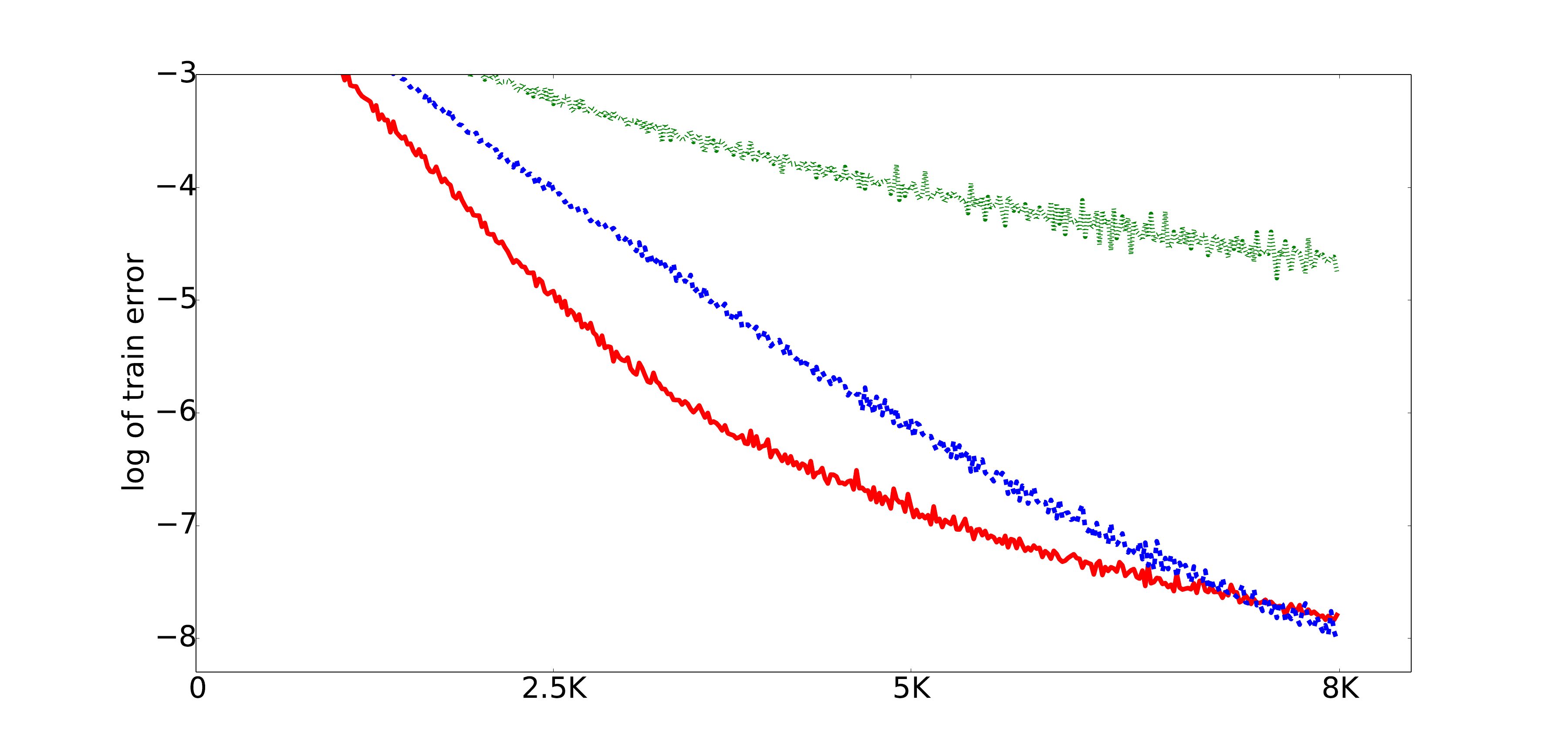}
\includegraphics[width=.5\textwidth]{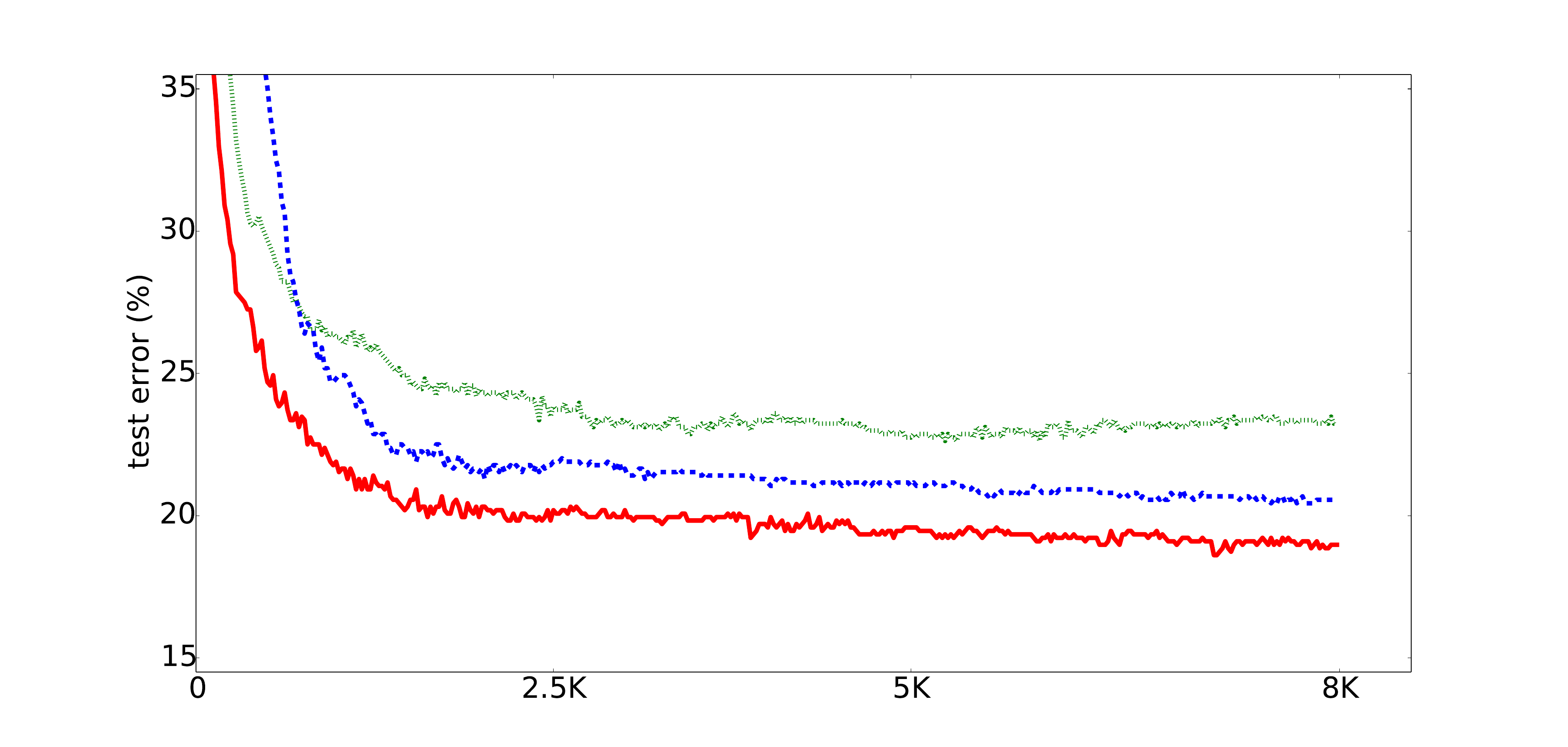}
\caption{
    The plot depicts the training error (left) on a log scale and test error
    (right) as percentage of misclassified examples for fold 4 of TFD.  On the
    x-axis we have the number of updates. Dotted green line (top, worst) stands
    for using the same minibatch (of 256 examples) to compute the gradient and
    evaluate the metric. Dashed blue line (middle) uses a different minibatch
    of 384 examples from the training set to evaluate the metric.  The solid
    red line (bottom, best) relies on a randomly sampled batch of unlabeled
    data to estimate the metric.
} 
\label{fig:overfitting}
\end{figure}

Statistically, if both expectations over $\mathbf{x}$ are computed from the
same samples, the two estimations are not independent from each other.  We
hypothesize that this can lead to the overfitting of the current minibatch.
Fig.~\ref{fig:overfitting} provides empirical evidence that our hypothesis is
correct. \citet{VinyalsP12} make a similar empirical observation.

As discussed in section \ref{sec:prop}, enforcing a constant change in the
model distribution helps ensuring stable progress but also protects from large
changes in the model which can be detrimental (could result in overfitting a
subset of the data). We get this effect as long as the metric provides a good
measure of how much the model distribution changes. Unfortunately the metric is
computed over training examples, and hence it will focus on how much $p$
changes at these points. When learning overfits the training set we usually
observe reduction in the training error that result in larger increases of the
generalization error.  This behaviour can be avoided by natural gradient
descent if we can measure how much $p$ changes far away from the training set.
To explore this idea we propose to increase the accuracy of the metric
$\mathbf{F}$ by using unlabeled data, helping us to measure how $p$ changes far
away from the training set.

We explore empirically these two hypotheses on the Toronto Face Dataset
(TFD)~\citep{Susskind2010} which has a small training set and a large pool of
unlabeled data.  Fig. \ref{fig:overfitting} shows the training and test error
of a model trained on fold 4 of TFD, though similar results are obtained for
the other folds.

We used a two layer model, where the first layer is convolutional. It uses 512
filters of 14X14, and applies a sigmoid activation function. The next layer is
a dense sigmoidal one with  1024 hidden units. The output layer uses sigmoids
as well instead of softmax.  The data is pre-processed by using local contrast
normalization.  Hyper-parameters have been selected using a grid search (more
details in the appendix).

We notice that {\em re-using the same samples for the metric and gradient
results in worse global training error} (training error over the entire train
set) {\em and worse test error}. 

Our intuition is that we are seeing the model overfitting, at each step, the
current training minibatch. At each step we compute the gradient and the metric
on the same example. There can be, within this  minibatch, some direction in
which we could overfit some specific property of these examples. Because we
only look at how the model changes at these points when computing the metric,
all gradients will agree with this direction. Specifically, the model believes
that moving in this direction will lead to steady change in the KL (the KL
looks at the covariance between the gradeints from the output layer to the
parameters, which will be low in the picked direction) and the model will
believe it can take a large step, learning this specific feature.  However it
is not useful for generalization, nor is it useful for the other training
examples (e.g. if the particular deformation is not actually correlated with
the classes it needs to predict). This means that on subsequent training
examples the model will under-perform, resulting in a worse overall error. 

On the other hand, if we use a different minibatch for the gradient and for the
metric, it is less likely that the same particular feature to be present in
both the set of examples used for the gradient and those used for the metric.
So either the gradient will not point in said direction (as the feature is not
present in the gradient), or, if the feature is present in the gradient, it may
not be present in the examples used for computing the metric. That would lead
to a larger variance, and hance the model is less likely to take a large step
in this direction.

Using unlabeled data results in better test error for worse training error,
which means that this way of using the unlabeled data acts like a regularizer.

\vspace*{-2mm}
\section{Robustness to reorderings of the train set}
\vspace*{-2mm}
\label{sec:robustness}

We want to test the hypothesis that, by forcing the change in the model, in
terms of the KL, to be constant, the model avoids to take large steps, being
more robust to the reordering of the train set.

We repeat the experiment from \cite{Erhan-aistats-2010-small}, using the NISTP
dataset introduced in \cite{Bengio+al-AI-2011-small} (which is just the NIST
dataset plus deformations) and use 32.7M samples of this data. The original
experiment attempts to measure the importance of the early examples in the
learnt model. To achieve this we respect the same protocol as the original
paper described below:

\begin{algorithm}
    \caption{Protocol for the robustness to the order of training examples}
    \begin{algorithmic}[1]
    \STATE Split the training data into two large chunks of 16.4M data points
    \STATE Split again the first chunk into 10 equal size segments
    \FOR {i between 1 and 10}
    \FOR {steps between 1 and 5}
            \STATE Replace segment i by new randomly sampled examples that have not been used before
            \STATE Train the model from scratch
            \STATE Evaluate the model on 10K heldout examples
        \ENDFOR
        \STATE Compute the segment i mean variance, among the 5 runs in the output 
        of the trained model (the variance in the activations of the output layer)
    \ENDFOR
\STATE Plot the mean variance as a function of which segment was resampled
\end{algorithmic}
\end{algorithm}

\begin{figure}
\begin{center}
\includegraphics[width=.7\linewidth]{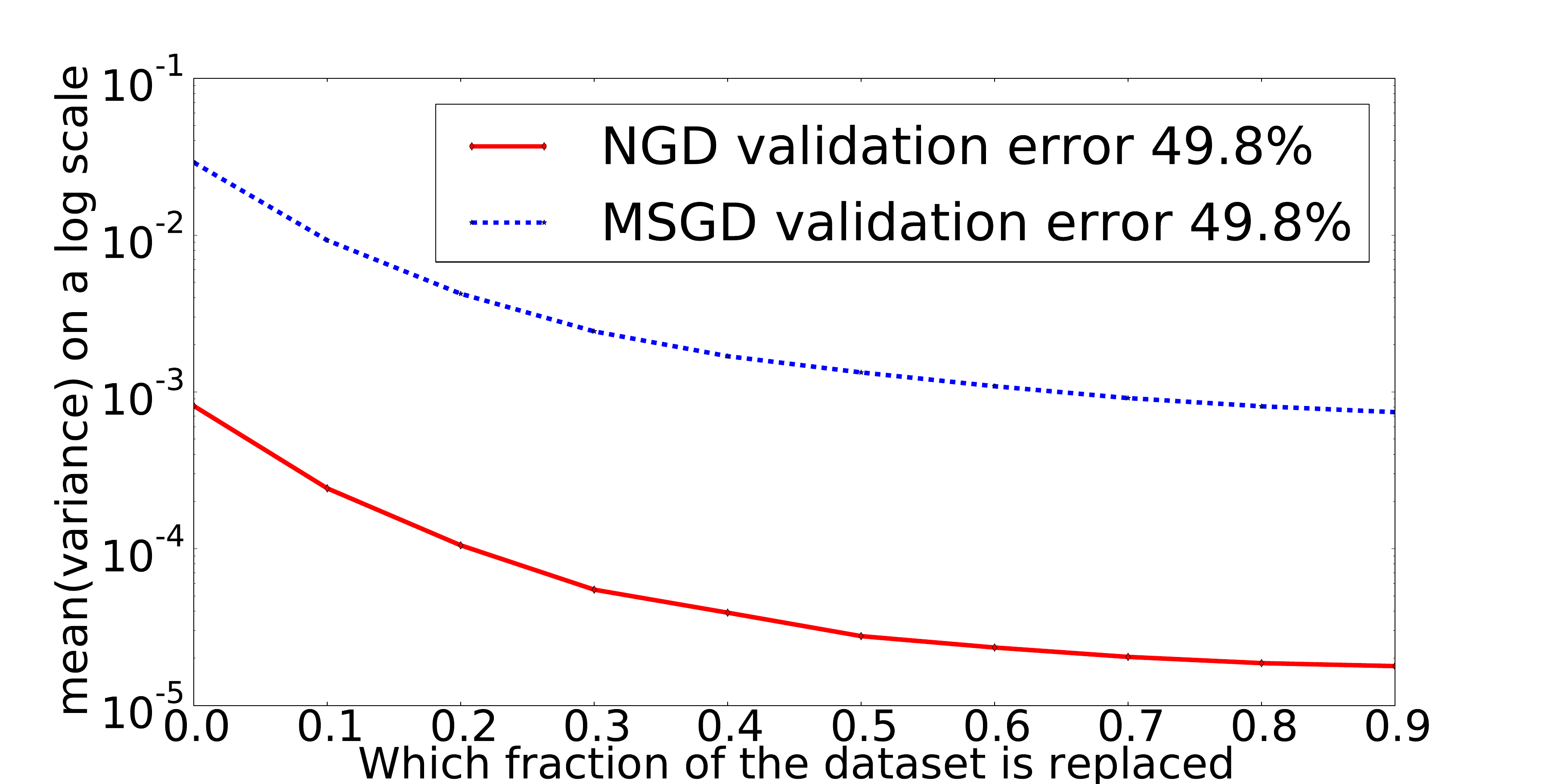}
\caption{
    The plot describes how much the model is influenced by different parts of
    an online training set. The variance induced by re-shuffling of data for
    natural gradient descent is order of magnitudes lower than for SGD. See
    text for more information.
} 
\label{fig:train_order}
\end{center}
\end{figure}

Fig.~\ref{fig:train_order} shows these curves for minibatch stochastic gradient
descent and natural gradient descent.

Note that the variance at each point on the curve depends on the speed with
which we move in functional space. For a fixed number of examples one can
artificially tweak the curves by decreasing the learning rate. With a smaller
learning rate we move slower, and hence the model, from a functional point of
view,  does not change by much, resulting in low variance. In the limit, with a
learning rate of 0, the model always stays the same.  In order to be fair to
the two algorithms compared in the plot, natural gradient descent and
stochastic gradient descent, we use the error on a different validation set as
a measure of how much we moved in the functional space. This helps us to chose
hyper-parameters such that after 32.7M samples both methods achieve about the
same validation error of 49.8\% (see appendix for hyper-parameters). Both
models are run on minibatches of the same size, 512 samples.  We use the same
minibatch to compute the metric as we do to compute the gradient for natural
gradient descent, to avoid favorazing NGD over MSGD.

The results are consistent with our hypothesis that natural gradient descent
avoids making large steps in function space during training, staying on the
path that induces least variance.  Note that relative variance might not be
affected when switching to natural gradient descent. That is to be expected,
as, in the initial stages of learning, any gradient descent algorithm needs to
choose a basin of attraction (a local minima) while subsequent steps make
progress towards these minima. That is, the first examples must have,
relatively speaking, a big impact on the learnt model. However what we are
trying to investigate is the overall variance. Natural gradient descent (in
global terms) has lower variance, resulting in more consistent functional
behaviour regardless of the order of the input examples.

SGD might move in a direction early on that could possibly yield overfitting
(e.g. getting forced into some quadrant of parameter space based only on a few
examples) resulting in different models at the end.  This suggests that natural
gradient descent can deal better with nonstationary data and can be less
sensitive to the particular examples selected early on during training.

\vspace*{-2mm}
\section{Natural conjugate gradient}
\vspace*{-2mm}
\label{sec:nncg}
Natural gradient descent is a first order method in the space of functions,
which raises the question: can the algorithm be improved by considering second
order information? Unfortunately computing the \emph{Hessian on the manifold}
can be daunting (especially from a computational perspective). That is the
Hessian of the error as a function of the density probability functions
$p_\theta(\mathbf{t}|\mathbf{x})$ that give rise to the manifold structure.
Note that in \cite{RouxF10} a method for combining second order information and
TONGA is proposed. It is not clear how this algorithm  can be applied to
natural gradient descent.  Also in \cite{RouxF10} a different Hessian matrix is
considered, namely that of the error as a function of the parameter. 

A simpler option is, however, to use an optimization method such as Nonlinear
Conjugate Gradient, that is able to take advantage of second order structure
(by following locally conjugate directions) while not requiring to actually
compute the Hessian at any point in time, only the gradient.
\citet{AbsMahSep2008} describes how second order methods can be generalized to
the manifold case.  In \citet{HonkelaTRK08, HonkelaRKTK10} a similar idea is
proposed in the context of variational inference and specific assumptions on
the form of $p_\theta$ are made.  \citet{GonzalezD06} is more similar in spirit
with this work, though their approach is defined for the diagonal form of the
Fisher Information Matrix and differs in how they propose to compute a new
conjugate direction. 

Natural conjugate gradient, the manifold version of nonlinear conjugate
gradient, is defined following the same intuitions~\citep{Shewchuck94}.  The
only problematic part of the original algorithm is how to obtain a new
conjugate direction given the previous search direction and the current natural
gradient.  The problem arises from the fact that the two vectors belong to
different spaces.  The local geometry around the point where the previous
search direction $d_{t-1}$ was computed is defined by $\mathbf{F}_{t-1}$, while
the geometry around the new direction is defined by $\mathbf{F}_{t}$, where, in
principle, $\mathbf{F}_{t-1} \neq \mathbf{F}_{t}$.
Following~\cite{AbsMahSep2008} we would need to ``transport'' $d_{t-1}$ into
the space of $\nabla_{N_t}$, an expensive operation, before we can compute a
new direction using a standard formula like Polak-Ribiere. Furthermore, the
line search should be done along the geodesic of the manifold. 

\citet{GonzalezD06, HonkelaRKTK10} address these issues by making the
assumption that $\mathbf{F}_{t-1}$ and $\mathbf{F}_t$ are identical (so
$d_{t-1}$ does not need to be transported). This assumption is detrimental to
the algorithm because it goes against what we want to achieve. By employing a
conjugate gradient method we hope to make large steps, from which it follows
that the metric is very likely to change. Hence the assumption can not hold.

Computing the right direction is difficult, as the transportation operator for
$d_{t-1}$ is hard to compute in a generic manner, without enforcing strict
constraints on the form of $p_{\theta}$. Inspired by our new interpretation of
the KSD algorithm, we propose instead to solve for the coefficients $\beta$ and
$\alpha$ that minimizes our cost, where $\frac{\beta}{\alpha}$ is the
coefficient required in computing the new direction and $\alpha$ is the step
size:
\begin{equation}
\min_{\alpha, \beta} \mathcal{L}\left(\theta_{t-1} +  \left[\begin{array}{l} \alpha_t \\ \beta_t \end{array}\right] \left[ \begin{array}{l}
\nabla_{N_t} \\ d_{t-1} \end{array} \right]\right)
\end{equation}
The new direction therefore is:
\begin{equation}
d_t = \nabla_{N_t} + \frac{\beta_t}{\alpha_t} d_{t-1}
\end{equation}

Note that the resulting algorithm looks identical to standard nonlinear
conjugate gradient. The difference are that we use the natural direction
$\nabla_{N_t}$ instead of the gradients $\nabla_t$ everywhere and that we use
an off the shelf solver to optimize simultaneously for both the step size
$\alpha_t$ and the correction term $\frac{\beta_t}{\alpha_t}$ used to compute
the new direction. 

We can show that in the Euclidean space, given a second order Taylor
approximation of $\mathcal{L}(\theta_{t-1})$, under the assumption that the
Hessian $\mathbf{H}_{t-1}$ is symmetric, this approach will result in conjugate
direction. Because we do not do the multidimensional line search along
geodesics or transport the gradients in the same space, in general, this
approach is also just an approximation to the locally conjugate direction on
the manifold.   

The advantage of our approach over using some formula, like Polak-Ribiere, and
ignoring the difference between $\mathbf{F}_{t-1}$ and $\mathbf{F}_t$, is that
we are guaranteed to minimize the cost $\mathcal{L}$ at at each step.

Let us show that the resulting direction is conjugate to the previous one in
Euclidean space.  Let $d_{t-1}$ be the previous direction and $\gamma_{t-1}$
the step size such that \mbox{$\mathcal{L}(\theta_{t-1} + \gamma_{t-1}
d_{t-1})$} is minimal for fixed $d_{t-1}$.  If we approximate $\mathcal{L}$ by
its second order Taylor expansion and compute the derivative with respect to
the step size $\gamma_{t-1}$ we have that:
\begin{equation}
\label{eq:prev}
\frac{\partial \mathcal{L}}{\partial \theta} d_{t-1}+ \gamma_{t-1} d_{t-1}^T \mathbf{H} d_{t-1} = 0 
\end{equation}
Suppose now that we take the next step which takes the form of
\begin{equation*}
\begin{array}{ll}
\mathcal{L}(\theta_t + \beta_t d_{t-1} + \alpha_t \nabla_{N_t}^T) &\\
& \hspace*{-3cm}= \mathcal{L}(\theta_{t-1} + \gamma_{t-1} d_{t-1} + \beta_{t} d_{t-1} + \alpha_t \nabla_{N_t}^T)
\end{array}
\end{equation*}

where we minimize for $\alpha_t$ and $\beta_t$. If we replace $\mathcal{L}$ by
its second order Taylor series around $\theta_{t-1}$, compute the derivative
with respect to $\beta_t$ and use the assumption that $\mathbf{H}_{t-1}$ (where
we drop the subscript) is symmetric, we get:

$\frac{\partial \mathcal{L}}{\partial \theta} d_{t-1} + \alpha_t \nabla_{N_t}  \mathbf{H} d_{t-1} + (\gamma_{t-1} + \beta_t) d_{t-1}^T \mathbf{H} d_{t-1} = 0$

Using the previous relation~\ref{eq:prev} this implies that

$(\alpha_t \nabla_{N_t} \mathcal{L}^T + \beta_t d_{t-1})^T \mathbf{H} d_{t-1} = 0$, i.e. that the new direction 
is conjugate to the last one.

\vspace*{-2mm}
\section{Benchmark}
\vspace*{-2mm}
\label{sec:benchmark}
We carry out a benchmark on the Curves dataset, using the 6 layer deep
auto-encoder from \citet{Martens10}.  The dataset is small, has only 20K
training examples of 784 dimensions each.  All methods except SGD use the
truncated Newton pipeline. The benchmark is run on a GTX 580 Nvidia card, using
Theano~\citep{bergstra+al:2010-scipy-short} for cuda kernels.  Fig.
\ref{fig:benchmark} contains the results. See supplementary materials for
details.

For natural gradient we present two runs, one in which we force the model to
use small minibatches of 5000 examples, and one in which we run in batch mode.
In both cases all other hyper-parameters were manually tuned.  The learning
rate is fixed at $1.0$ for the model using a minibatch size of 5000 and a line
search was used when running in batch mode.  For natural conjugate gradient we
use {\tt scipy.optimize.fmin\_cobyla} to solve the two dimensional line search.
This corresponds to the new algorithm introduced in section \ref{sec:nncg}.  We
additionally run the same algorithm, where we rely on the Polak-Riebiere
formula for computing the new direction and  use standard line search for the
learning rate. This corresponds to our implementation of the algorithm proposed
in \citet{GonzalezD06}, where instead of a diagonal approximation of the metric
we reply on a truncated Newton approach.  In both cases we run in full batch
mode.  For SGD we used a minibatch of 100 examples and a learning rate of
$0.01$.  Note that our implementation of natural gradient does not directly
matches James Martens' Hessian Free as there are components of the algorithm
that we do not use like backtracking and preconditioning.

\begin{figure}[ht] 
    \includegraphics[width=0.53\textwidth]{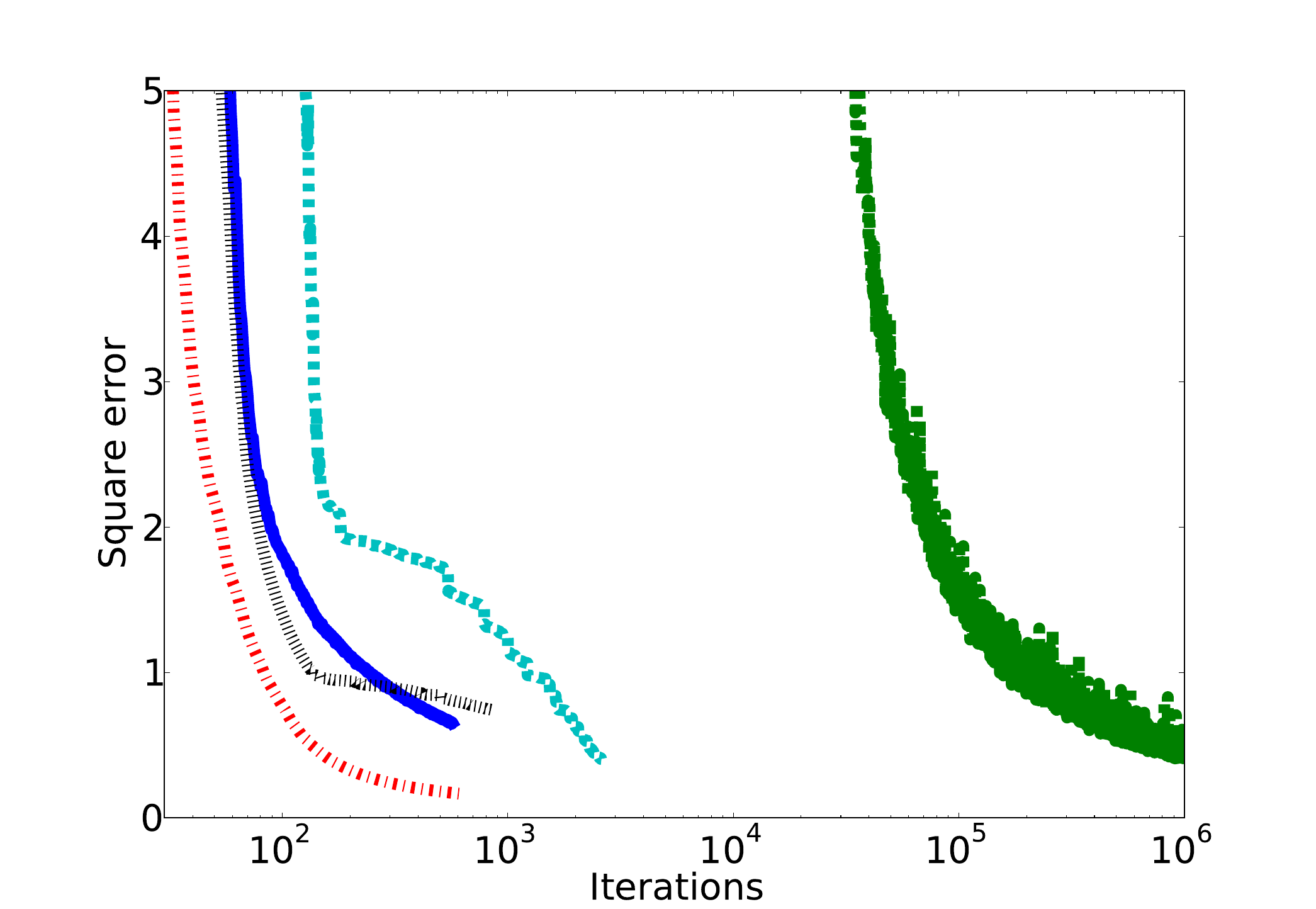}
    \includegraphics[width=.53\textwidth]{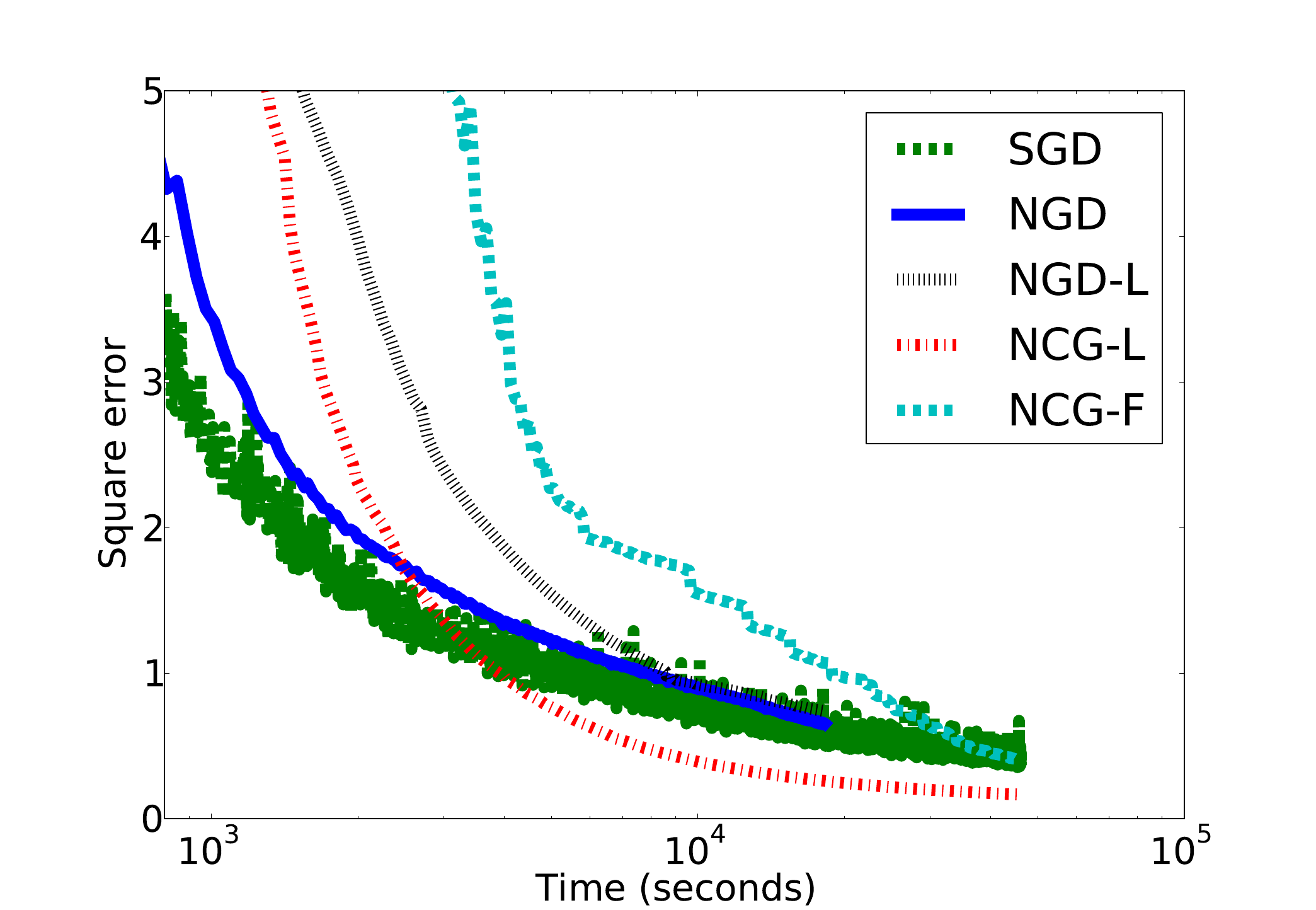} 
    \caption{
        The plots depict the training curves on a log scale (x-axis) for
        different learning algorithms on the Curves dataset. On the y-axis we
        have training error on the whole dataset.  The left plot shows the
        curves in terms of iterations, while the right plot show the curves as
        a function of clock time (seconds). NGD stands for natural gradient
        descent descent with a minibatch of 1000 and a fixed learning rate.
        NGD-L is natural gradient descent in batch mode where we use a line
        search for the learning rate.  NCG-L uses an {\tt
        scipy.optimize.fmin\_cobyla} for finding both learning rate and the new
        conjugate direction. NCG-F employs the Polak-Ribiere formula, ignoring
        the change in the metric. SGD stands for stochastic gradient descent.  
    } 
    \label{fig:benchmark} 
\end{figure}

The first observation that we can make is that natural gradient descent can run
reliably with smaller minibatches, as long as the gradient and the metric are
computed on different samples and the learning rate used is sufficiently small
to account for the noise on the natural direction (or the damping coefficient 
is sufficiently large).  This can be seen from
comparing the two runs of natural gradient descent descent.  Our result is in
the spirit of the work of \cite{Kiros}, though the exact approach of dealing
with small minibatches is slightly different.

The second observation that we can draw is that incorporating second order
information into natural gradient descent might be beneficial. We see a speedup
in convergence. The result agrees with the observation made in
\cite{VinyalsP12}, where Krylov Subspace Descent (KSD) was shown to converge slightly
faster than Hessian-Free Optimization.  Given the relationship between natural
gradient and Hessian-Free Optimization, and between KSD and natural conjugate
gradient, we expected to see the same trend between natural gradient descent
and natural conjugate gradient. We regard our algorithm as similar to KSD, with
its main advantage being that it avoids storing in memory the Krylov subspace.
Furthermore, from the comparison between NCG-L and NCG-F we can see evidence
for our argumentation that it is useful to use an optimizer for finding the new
direction.  The reason why NCG-F under performs is that the smooth change in
the metric assumption contradicts what the algorithm is trying to do, namely
move as far as possible in parameter space.

\vspace*{-2mm}
\section{Discussion and conclusions}
\vspace*{-2mm}

In this paper we have made the following original contributions. We showed that
by employing the extended Gauss-Newton approximation of the Hessian both
Hessian-Free Optimization and Krylov Subspace Descent can be interpreted as
implementing natural gradient descent. Furthermore, by adding the previous
search direction to the Krylov subspace, KSD does something akin to an
approximation to nonlinear conjugate gradient on the manifold. 

We proposed an approximation of nonlinear conjugate gradient on the manifold
which is similar in spirit to KSD, with the difference that we rely on linear
conjugate gradient to invert the metric $\mathbf{F}$ instead of using a Krylov
subspace.  This allows us to save on the memory requirements which can become
prohibitive for large models, while still retaining some of the second order
information of the cost and hence converging faster than vanilla natural
gradient. 

Lastly we highlighted the difference between natural gradient descent and
TONGA, and brought forward two new properties of the algorithm compared to
stochastic gradient descent. The first one is an empirical evaluation of the
robustness of natural gradient descent to the order of examples in the training
set, robustness that can be crucial when dealing with non-stationary data. The
second property is the ability of natural gradient descent to guard against
large drops in generalization error especially when the accuracy of the metric
is increased by using unlabeled data. We believe that these properties as well
as others found in the literature and shortly summarized in section
\ref{sec:prop} might provide some insight in the recent success of Hessian Free
and KSD for deep models.

Finally we make the empirical observation that natural gradient descent can
perform well even when its metric and gradients are estimated on rather small
minibatches. One just needs to use different samples for computing the
gradients from those used for the metric. Also the step size and damping
coefficient need to be bounded to account for the noise in the computed descent
direction.

\subsubsection*{Acknowledgments}
We would like to thank the developers of
Theano~\citep{bergstra+al:2010-scipy,Bastien-Theano-2012} and Guillaume Desjardins and Yann Dauphin for 
their insightful comments.  We would also like
to thank NSERC, Compute Canada, and Calcul Qu\'ebec for providing computational
resources. Razvan Pascanu is supported by a DeepMind Fellowship.  

{\small
\bibliography{strings,strings-short,strings-shorter,aigaion-shorter,ml,myrefs}

\begin{thebibliography}{}

\bibitem[Absil {\em et~al.}(2008)Absil, Mahony, and Sepulchre]{AbsMahSep2008}
Absil, P.-A., Mahony, R., and Sepulchre, R. (2008).
\newblock {\em Optimization Algorithms on Matrix Manifolds\/}.
\newblock Princeton University Press, Princeton, NJ.

\bibitem[Amari(1985)Amari]{Amari-1985}
Amari, S. (1985).
\newblock Differential geometrical methods in statistics.
\newblock {\em Lecture notes in statistics\/}, {\bf 28}.

\bibitem[Amari(1997)Amari]{Amari97}
Amari, S. (1997).
\newblock Neural learning in structured parameter spaces - natural {R}iemannian
  gradient.
\newblock In {\em In Advances in Neural Information Processing Systems\/},
  pages 127--133. MIT Press.

\bibitem[Amari {\em et~al.}(1992)Amari, Kurata, and Nagaoka]{Amari-1992}
Amari, S., Kurata, K., and Nagaoka, H. (1992).
\newblock Information geometry of {B}oltzmann machines.
\newblock {\em {IEEE} Trans. on Neural Networks\/}, {\bf 3}, 260--271.

\bibitem[Amari(1998)Amari]{Amari98}
Amari, S.-I. (1998).
\newblock Natural gradient works efficiently in learning.
\newblock {\em Neural Comput.}, {\bf 10}(2), 251--276.

\bibitem[Arnold {\em et~al.}(2011)Arnold, Auger, Hansen, and Olivier]{Arnold11}
Arnold, L., Auger, A., Hansen, N., and Olivier, Y. (2011).
\newblock Information-geometric optimization algorithms: A unifying picture via
  invariance principles.
\newblock {\em CoRR\/}, {\bf abs/1106.3708}.

\bibitem[Bastien {\em et~al.}(2012)Bastien, Lamblin, Pascanu, Bergstra,
  Goodfellow, Bergeron, Bouchard, and Bengio]{Bastien-Theano-2012}
Bastien, F., Lamblin, P., Pascanu, R., Bergstra, J., Goodfellow, I., Bergeron,
  A., Bouchard, N., and Bengio, Y. (2012).
\newblock Theano: new features and speed improvements.
\newblock Submited to Deep Learning and Unsupervised Feature Learning NIPS 2012
  Workshop.

\bibitem[Bengio {\em et~al.}(2011)Bengio, Bastien, Bergeron,
  Boulanger-Lewandowski, Breuel, Chherawala, Cisse, C{\^{o}}t{\'{e}}, Erhan,
  Eustache, Glorot, Muller, Pannetier~Lebeuf, Pascanu, Rifai, Savard, and
  Sicard]{Bengio+al-AI-2011-small}
Bengio, Y., Bastien, F., Bergeron, A., Boulanger-Lewandowski, N., Breuel, T.,
  Chherawala, Y., Cisse, M., C{\^{o}}t{\'{e}}, M., Erhan, D., Eustache, J.,
  Glorot, X., Muller, X., Pannetier~Lebeuf, S., Pascanu, R., Rifai, S., Savard,
  F., and Sicard, G. (2011).
\newblock Deep learners benefit more from out-of-distribution examples.
\newblock In {\em JMLR W\&CP: Proc. AISTATS'2011\/}.

\bibitem[Bergstra {\em et~al.}(2010a)Bergstra, Breuleux, Bastien, Lamblin,
  Pascanu, Desjardins, Turian, Warde-Farley, and
  Bengio]{bergstra+al:2010-scipy-short}
Bergstra, J., Breuleux, O., Bastien, F., Lamblin, P., Pascanu, R., Desjardins,
  G., Turian, J., Warde-Farley, D., and Bengio, Y. (2010a).
\newblock Theano: a {CPU} and {GPU} math expression compiler.
\newblock In {\em Proceedings of the Python for Scientific Computing Conference
  ({SciPy})\/}.

\bibitem[Bergstra {\em et~al.}(2010b)Bergstra, Breuleux, Bastien, Lamblin,
  Pascanu, Desjardins, Turian, Warde-Farley, and
  Bengio]{bergstra+al:2010-scipy}
Bergstra, J., Breuleux, O., Bastien, F., Lamblin, P., Pascanu, R., Desjardins,
  G., Turian, J., Warde-Farley, D., and Bengio, Y. (2010b).
\newblock Theano: a {CPU} and {GPU} math expression compiler.
\newblock In {\em Proceedings of the Python for Scientific Computing Conference
  ({SciPy})\/}.
\newblock Oral Presentation.

\bibitem[Bishop(2006)Bishop]{Bishop}
Bishop, C.~M. (2006).
\newblock {\em Pattern Recognition and Machine Learning (Information Science
  and Statistics)\/}.
\newblock Springer-Verlag New York, Inc.

\bibitem[Chapelle and Erhan(2011)Chapelle and Erhan]{chapelle11}
Chapelle, O. and Erhan, D. (2011).
\newblock {Improved Preconditioner for Hessian Free Optimization}.
\newblock In {\em NIPS Workshop on Deep Learning and Unsupervised Feature
  Learning\/}.

\bibitem[Choi {\em et~al.}(2011)Choi, Paige, and Saunders]{CPS11}
Choi, S.-C.~T., Paige, C.~C., and Saunders, M.~A. (2011).
\newblock {MINRES-QLP}: {A} {K}rylov subspace method for indefinite or singular
  symmetric systems.
\newblock {\bf 33}(4), 1810--1836.

\bibitem[Desjardins {\em et~al.}(2013)Desjardins, Pascanu, Courville, and
  Bengio]{DesjardinsPascanu13}
Desjardins, G., Pascanu, R., Courville, A., and Bengio, Y. (2013).
\newblock Metric-free natural gradient for joint-training of boltzmann
  machines.
\newblock {\em ICLR\/}, {\bf abs/1301.3545}.

\bibitem[Erhan {\em et~al.}(2010)Erhan, Courville, Bengio, and
  Vincent]{Erhan-aistats-2010-small}
Erhan, D., Courville, A., Bengio, Y., and Vincent, P. (2010).
\newblock Why does unsupervised pre-training help deep learning?
\newblock In {\em JMLR W\&CP: Proc. AISTATS'2010\/}, volume~9, pages 201--208.

\bibitem[Gonzalez and Dorronsoro(2006)Gonzalez and Dorronsoro]{GonzalezD06}
Gonzalez, A. and Dorronsoro, J. (2006).
\newblock Natural conjugate gradient training of multilayer perceptrons.
\newblock {\em Artificial Neural Networks – ICANN 2006\/}, pages 169--177.

\bibitem[Heskes(2000)Heskes]{hes99b}
Heskes, T. (2000).
\newblock On natural learning and pruning in multilayered perceptrons.
\newblock {\em Neural Computation\/}, {\bf 12}, 1037--1057.

\bibitem[Honkela {\em et~al.}(2008)Honkela, Tornio, Raiko, and
  Karhunen]{HonkelaTRK08}
Honkela, A., Tornio, M., Raiko, T., and Karhunen, J. (2008).
\newblock Natural conjugate gradient in variational inference.
\newblock In {\em Neural Information Processing\/}, pages 305--314.

\bibitem[Honkela {\em et~al.}(2010)Honkela, Raiko, Kuusela, Tornio, and
  Karhunen]{HonkelaRKTK10}
Honkela, A., Raiko, T., Kuusela, M., Tornio, M., and Karhunen, J. (2010).
\newblock Approximate riemannian conjugate gradient learning for fixed-form
  variational bayes.
\newblock {\em Journal of Machine Learning Research\/}, {\bf 11}, 3235--3268.

\bibitem[Kakade(2001)Kakade]{Kakade01}
Kakade, S. (2001).
\newblock A natural policy gradient.
\newblock In {\em NIPS\/}, pages 1531--1538. MIT Press.

\bibitem[Kiros(2013)Kiros]{Kiros}
Kiros, R. (2013).
\newblock Training neural networks with stochastic hessian-free optimization.
\newblock {\em ICLR\/}.

\bibitem[{Le Roux} {\em et~al.}(2008){Le Roux}, Manzagol, and
  Bengio]{LeRoux+al-tonga-2008-small}
{Le Roux}, N., Manzagol, P.-A., and Bengio, Y. (2008).
\newblock Topmoumoute online natural gradient algorithm.
\newblock In {\em {NIPS'07}\/}.

\bibitem[Martens(2010)Martens]{Martens10}
Martens, J. (2010).
\newblock Deep learning via hessian-free optimization.
\newblock In {\em ICML\/}, pages 735--742.

\bibitem[Martens and Sutskever(2011)Martens and Sutskever]{Martens11}
Martens, J. and Sutskever, I. (2011).
\newblock Learning recurrent neural networks with hessian-free optimization.
\newblock In {\em ICML\/}, pages 1017--1024.

\bibitem[Mizutani and Demmel(2003)Mizutani and Demmel]{Mizutani}
Mizutani, E. and Demmel, J. (2003).
\newblock Iterative scaled trust-region learning in krylov subspaces via
  peralmutter's implicit sparse hessian-vector multiply.
\newblock In {\em NIPS\/}, pages 209--216.

\bibitem[Nocedal and Wright(2000)Nocedal and Wright]{nocedal99}
Nocedal, J. and Wright, S.~J. (2000).
\newblock {\em {Numerical Optimization}\/}.
\newblock Springer.

\bibitem[Park {\em et~al.}(2000)Park, Amari, and Fukumizu]{Park00}
Park, H., Amari, S.-I., and Fukumizu, K. (2000).
\newblock Adaptive natural gradient learning algorithms for various stochastic
  models.
\newblock {\em Neural Networks\/}, {\bf 13}(7), 755 -- 764.

\bibitem[Pearlmutter(1994)Pearlmutter]{Pearlmutter94}
Pearlmutter, B.~A. (1994).
\newblock Fast exact multiplication by the hessian.
\newblock {\em Neural Computation\/}, {\bf 6}, 147--160.

\bibitem[Peters and Schaal(2008)Peters and Schaal]{Peters_N_2008}
Peters, J. and Schaal, S. (2008).
\newblock Natural actor-critic.
\newblock (7-9), 1180--1190.

\bibitem[Roux and Fitzgibbon(2010)Roux and Fitzgibbon]{RouxF10}
Roux, N.~L. and Fitzgibbon, A.~W. (2010).
\newblock A fast natural newton method.
\newblock In {\em ICML\/}, pages 623--630.

\bibitem[Schaul(2012)Schaul]{Schaul2012proof}
Schaul, T. (2012).
\newblock Natural evolution strategies converge on sphere functions.
\newblock In {\em Genetic and Evolutionary Computation Conference (GECCO)\/}.

\bibitem[Schraudolph(2002)Schraudolph]{Schraudolph02}
Schraudolph, N.~N. (2002).
\newblock Fast curvature matrix-vector products for second-order gradient
  descent.
\newblock {\em Neural Computation\/}, {\bf 14}(7), 1723--1738.

\bibitem[Shewchuck(1994)Shewchuck]{Shewchuck94}
Shewchuck, J. (1994).
\newblock An introduction to the conjugate gradient method without the
  agonizing pain.
\newblock Technical report, CMU.

\bibitem[Sohl-Dickstein(2012)Sohl-Dickstein]{Dickstein12}
Sohl-Dickstein, J. (2012).
\newblock The natural gradient by analogy to signal whitening, and recipes and
  tricks for its use.
\newblock {\em CoRR\/}, {\bf abs/1205.1828}.

\bibitem[Sun {\em et~al.}(2009)Sun, Wierstra, Schaul, and Schmidhuber]{YiWSS09}
Sun, Y., Wierstra, D., Schaul, T., and Schmidhuber, J. (2009).
\newblock Stochastic search using the natural gradient.
\newblock In {\em ICML\/}, page 146.

\bibitem[Susskind {\em et~al.}(2010)Susskind, Anderson, and
  Hinton]{Susskind2010}
Susskind, J., Anderson, A., and Hinton, G.~E. (2010).
\newblock The {T}oronto face dataset.
\newblock Technical Report UTML TR 2010-001, U. Toronto.

\bibitem[Vinyals and Povey(2012)Vinyals and Povey]{VinyalsP12}
Vinyals, O. and Povey, D. (2012).
\newblock {Krylov Subspace Descent for Deep Learning}.
\newblock In {\em AISTATS\/}.

\end{thebibliography}
\bibliographystyle{natbib}
}

\section*{ Appendix}
\subsection*{Expected Hessian to Fisher Information Matrix}

 The Fisher Information Matrix form can be obtained from the 
expected value of the Hessian :  
\begin{eqnarray}
    \label{eq:relation1}
\mathbb{E}_{\mathbf{z}}\left[-\frac{\partial^2 \log p_{\theta}}{\partial \theta}\right] &= &
  \mathbb{E}_{\mathbf{z}}\left[-\frac{\partial \frac{1}{p_{\theta}}\frac{\partial p_{\theta}}{\partial \theta}}{\partial \theta} \right] =
  \mathbb{E}_{\mathbf{z}}
  \left[-\frac{1}{p_{\theta}(\sample)}\frac{\partial^2 p_{\theta}}{\partial \theta^2}+
  \left(\frac{1}{p_{\theta}} \frac{\partial p_{\theta}}{\partial \theta} \right)^T 
  \left(\frac{1}{p_{\theta}} \frac{\partial p_{\theta}}{\partial \theta} \right)\right] \nonumber \\
  &=& -\frac{\partial^2}{\partial \theta^2}\left(\sum_{\mathbf{z}} p_{\theta}(\mathbf{z})\right) + 
  \mathbb{E}_{\mathbf{z}}\left[\left(\frac{\partial \log p_{\theta}(\mathbf{z})}{\partial \theta}\right)^T
  \left(\frac{\partial \log p_{\theta}(\mathbf{z})}{\partial \theta}\right) \right] \nonumber \\
  &=&   \mathbb{E}_{\mathbf{z}}\left[\left(\frac{\partial \log p_{\theta}(\mathbf{z})}{\partial \theta}\right)^T
  \left(\frac{\partial \log p_{\theta}(\mathbf{z})}{\partial \theta}\right) \right]
\end{eqnarray}

\subsection*{Derivation of the natural gradient descent metrics}

\subsubsection*{Linear activation function}

In the case of linear outputs we assume that each entry of the vector $t$,
$t_i$ comes from a Gaussian distribution centered around $\out_i(\mathbf{x})$
with some standard deviation $\beta$. From this it follows that:

\begin{equation}
\label{eq:prob_linear}
p_{\theta}(\mathbf{t} | \mathbf{x}) = \prod_{i=1}^o \mathcal{N}(t_i|\out(\mathbf{x}, \theta)_i, \beta^2)
\end{equation}

\begin{equation}
\label{eq:deriv_FIM_linear}
\begin{array}{lll}
\metric & = & \mathbb{E}_{\mathbf{x} \sim \tilde{q}} \left[
  \mathbb{E}_{\mathbf{t} \sim \mathcal{N}(\mathbf{t}|\out(\mathbf{x}, \theta), \beta^2\mathbf{I})}
   \left[
       \sum_{i=1}^o \left(\frac{\partial \log _{\theta}p(t_i| \out(\mathbf{x})_i}{\partial \theta}\right)^T
       \left(\frac{\partial \log p_{\theta}(t_i | \out(\mathbf{x})_i}{\partial \theta}\right) 
\right] \right] \\
& = & \mathbb{E}_{\mathbf{x} \sim \tilde{q}} \left[
  \sum_{i=1}^o \left[
   \mathbb{E}_{\mathbf{t} \sim \mathcal{N}(\mathbf{t}|\out(\mathbf{x}, \theta), \beta^2\mathbf{I})}
\left[
    \left(\frac{\partial (t_i - y_i)^2}{\partial \theta}\right)^T
    \left(\frac{\partial (t_i - y_i)^2}{\partial \theta}\right) \right] \right] \right] \\
& = & \mathbb{E}_{\mathbf{x} \sim \tilde{q}} \left[
    \sum_{i=1}^o \left[
    \mathbb{E}_{\mathbf{t} \sim \mathcal{N}(\mathbf{t}|\out(\mathbf{x}, \theta), \beta^2\mathbf{I})} \left[
      (t_i - y_i)^2 \left(\frac{\partial y_i}{\partial \theta}\right)^T
                    \left(\frac{\partial y_i}{\partial \theta}\right) \right] \right] \right] \\
& = & \mathbb{E}_{\mathbf{x} \sim \tilde{q}} \left[
    \sum_{i=1}^o \left[
    \mathbb{E}_{\mathbf{t} \sim \mathcal{N}(\mathbf{t}|\out(\mathbf{x}, \theta), \beta^\mathbf{I})}
\left[
      (t_i - y_i)^2\right] \left(\frac{\partial y_i}{\partial \theta}\right)^T
                    \left(\frac{\partial y_i}{\partial \theta}\right) \right] \right] \\
& = & \mathbb{E}_{\mathbf{x} \sim \tilde{q}} \left[
    \sum_{i=1}^o \left[
    \mathbb{E}_{\mathbf{t} \sim \mathcal{N}(\mathbf{t}|\out(\mathbf{x}, \theta), \beta^2\mathbf{I})}
\left[
      (t_i - y_i)^2\right] \left(\frac{\partial y_i}{\partial \theta}\right)^T
                    \left(\frac{\partial y_i}{\partial \theta}\right) \right] \right] \\
& = & \mathbb{E}_{\mathbf{x} \sim \tilde{q}} \left[
    \sum_{i=1}^o \left[
    \mathbb{E}_{\mathbf{t} \sim \mathcal{N}(\mathbf{t}|\out(\mathbf{x}, \theta), \beta^2\mathbf{I})}
 \left[
      (t_i - y_i)^2\right] \left(\frac{\partial y_i}{\partial \theta}\right)^T
                    \left(\frac{\partial y_i}{\partial \theta}\right) \right] \right] \\
& = & \beta^2 \mathbb{E}_{\mathbf{x} \sim \tilde{q}} \left[
    \sum_{i=1}^o \left[
     \left(\frac{\partial y_i}{\partial \theta}\right)^T
                    \left(\frac{\partial y_i}{\partial \theta}\right) \right] \right] \\
& = & \beta^2 \mathbb{E}_{\mathbf{x} \sim \tilde{q}} \left[ \jacob_{\out}^T \jacob_{\out} \right] \\
\end{array}
\end{equation}

\subsubsection*{Sigmoid activation function}

In the case of the sigmoid units, i,e, $\out = \text{sigmoid}(\lout)$,  we
assume a binomial distribution which gives us:

\begin{equation}
p(\mathbf{t}|\mathbf{x}) = \prod_i \out_i^{t_i}(1-\out_i)^{1-t_i}
\end{equation}

$\log p$ gives us the usual cross-entropy error used with sigmoid units. We can
compute the Fisher information matrix as follows:

\begin{equation}
\begin{array}{lll}
\metric & = & \mathbb{E}_{\mathbf{x} \sim \tilde{q}} \left[
\mathbb{E}_{\mathbf{t} \sim p(\mathbf{t}|\mathbf{x})} \left[
\sum_{i=1}^{o} \frac{(t_i-\outi_i)^2}{\outi_i^2 (1-\outi_i)^2} 
\left( \frac{\partial \outi_i}{\partial \theta}\right)^T \frac{\partial \outi_i}{\partial \theta}
\right] \right] \\
& = & \mathbb{E}_{\mathbf{x} \sim \tilde{q}} \left[
\sum_{i=1}^o \frac{1}{\outi_i(1-\outi_i)}
\left( \frac{\partial \outi_i}{\partial \theta}\right)^T \frac{\partial \outi_i}{\partial \theta}
\right] \\
& = & \mathbb{E}_{\mathbf{x} \sim \tilde{q}} \left[
\jacob_{\out}^T diag\left(\frac{1}{\out (1-\out)}\right) \jacob_{\out} \right]
\end{array}
\end{equation}

Note that $diag(\mathbf{v})$ stands for the diagonal matrix constructed from
the values of the vector $\mathbf{v}$ and we make an abuse of notation, where
by $\frac{1}{\mathbf{y}}$ we understand the vector obtain by applying the
division element-wise (the $i$-th element of is $\frac{1}{\mathbf{y}_i}$).

\subsubsection*{Softmax activation function}

For the softmax activation function, $\out=\text{softmax}(\lout)$, $p(\mathbf{t}|\mathbf{x})$ takes the form 
of a multinoulli:

\begin{equation}
    p(\mathbf{t}|\mathbf{x}) = \prod_{i=1}^o y_i^{t_i}
\end{equation}

\begin{equation}
\metric  =  \mathbf{E}_{\mathbf{x} \sim \tilde{q}} \left[
    \sum_{i=1}^o \frac{1}{\out_i} 
\left( \frac{\partial \outi_i}{\partial \theta}\right)^T \frac{\partial \outi_i}{\partial \theta}
\right] =  
\mathbf{E}_{\mathbf{x} \sim \tilde{q}} \left[
\jacob_{\out}^T diag\left(\frac{1}{\out}\right) \jacob_{\out} \right]
\end{equation}

\subsection*{Implementation Details}

We have implemented natural gradient descent descent using a truncated Newton
approach similar to the pipeline proposed by \cite{Pearlmutter94} and used by
\cite{Martens10}.  In order to better deal with singular and ill-conditioned
matrices we use the MinRes-QLP algorithm \citep{CPS11} instead of linear
conjugate gradient for certain experiments.  Both Minres-QLP as well as linear
conjugate gradient can be found implemented in Theano at
\href{https://github.com/pascanur/theano_optimize}{https://github.com/pascanur/theano\_optimize}.
We used the Theano library \citep{bergstra+al:2010-scipy-short} which allows
for a flexible implementation of the pipeline, that can automatically generate
the computational graph of the metric times some vector for different models:

\newcommand\codeComment[1]{\textcolor[rgb]{.3,.3,.3}{\textbf{#1}}}
\begin{Verbatim}[commandchars=\\\{\}]
import theano.tensor as TT
\codeComment{# `params` is the list of Theano variables containing the parameters}
\codeComment{# `vs` is the list of Theano variable representing the vector `v`}
\codeComment{# with whom we want to multiply the metric}
\codeComment{# `Gvs` is the list of Theano expressions representing the product} 
\codeComment{# between the metric and `vs`}

\codeComment{# `out_smx` is the output of the model with softmax units}
Gvs = TT.Lop(out_smx,params, 
             TT.Rop(out_smx,params,vs)/(out_smx*out_smx.shape[0]))

\codeComment{# `out_sig` is the output of the model with sigmoid units}
Gvs = TT.Lop(out_sig,params, 
             TT.Rop(out_sig,params,vs)/(out_sig*
                                        (1-out_sig)*
                                        out_sig.shape[0]))

\codeComment{# `out` is the output of the model with linear units}
Gvs = TT.Lop(out,params,TT.Rop(out,params,vs)/out.shape[0])
\end{Verbatim}

The full pseudo-code of the algorithm (which is very similar to the one for Hessian-Free Optimization) is 
given below. 
%The full Theano implementation can be retrieved from 
%\href{https://anon/anon}{https://anon/anon}.

\begin{algorithm}
\caption{Pseudocode for natural gradient descent algorithm}
\begin{algorithmic}[1]

    %\Comment{$gfn$ is a function that computes the metric times a vector}
    \STATE \codeComment{\# `gfn` is a function that computes the metric times some vector}
    \STATE gfn $\gets \left(\text{lambda } v \to \mathbf{F}v\right)$
    \WHILE{not \text{early\_stopping\_condition}}
    \STATE g $\gets \frac{\partial \mathcal{L}}{\partial \theta}$
    \STATE \codeComment{\# linear\_cg solves the linear system $G x = \frac{\partial \mathcal{L}}{\partial \theta}$}
    \STATE ng$ \gets$ linear\_cg(gfn, g, max\_iters = 20, rtol=1e-4)
    %\COMMENT{$\gamme$ is the learning rate}
    \STATE \codeComment{\# $\gamma$ is the learning rate}
    \STATE $\theta \gets \theta - \gamma$ ng
    \ENDWHILE
\end{algorithmic}
\end{algorithm}

Note that we do not usually do a line search for each step. While it could be
useful, conceptually we like to think of the algorithm as a first order method.
Doing a line search has the side effect that can lead to overfitting of the
current minibatch. To a certain extent this can be fixed by using new samples
of data to compute the error, though we find that we need to use reasonably
large batches to get an accurate measure. Using a learning rate as for a first
order method is a good alternative if one wants to apply this algorithm using
small minibatches of data. 

Even though we are ensured that $\metric$ is positive semi-definite by
construction, and MinRes-QLP is able to find a suitable solutions in case of
singular matrices, we still use a damping strategy for two reasons. The first
one is that we want to take in consideration the inaccuracy of the metric
(which is approximated only over a small minibatch).  The second reason is that
natural gradient descent makes sense only in the vicinity of $\theta$ as it is
obtained by using a Taylor series approximation, hence (as for ordinary second
order methods) it is appropriate to enforce a trust region for the gradient.
See \citet{Schaul2012proof}, where the convergence properties of natural
gradient descent (in a specific case) are studied. 

Following the functional manifold interpretation of the algorithm, we can
recover the Levenberg-Marquardt heuristic used in \citet{Martens10}.  We
consider only a first order Taylor approximation, where, due to the functional
manifold, for any function $f$,

\begin{equation}
f\left(\theta_t - \eta \mathbf{F}^{-1}\frac{\partial f(\theta_t)}{\partial \theta_t}^T\right) 
\approx f(\theta_t) - \eta \frac{\partial f(\theta_t)}{\partial \theta_t}
 \mathbf{F}^{-1} \frac{\partial f(\theta_t)}{\partial \theta_t}^T
\end{equation}

We now can check how well this approximation predicts the change in the error
for the step taken during an iteration. Based on this comparison we can define
our reduction ratio given by equation \eqref{eq:our_rho}. Based on this ratio we can
decide if we want to increase or decrease the damping applied to the metric
before inverting. As we will  show below, this ratio behaves identically with
the one in \citet{Martens10}.
\begin{equation}
\label{eq:our_rho}
\rho = \frac{f\left(\theta_t - \eta \mathbf{F}^{-1} \frac{\partial f(\theta_t)}{\partial \theta_t}^T\right) -  f(\theta_t)}
{- \eta \frac{\partial f(\theta_t)}{\partial \theta_t} \mathbf{F}^{-1} \frac{\partial f(\theta_t)}{\partial \theta_t}^T}
\end{equation}
%Let us massage the formula of $\rho$ coming from the second order
%approximation (Levenberg-Marquardt) used by~\citet{Martens10}, considering the
%step size $\Delta \theta$ to be in his case the inverse Hessian times the
%gradient, replacing the Hessian in this step by $\mathbf{F}$: \begin{equation}
%\label{eq:martens} \rho = \frac{f\left(\theta_t - \mathbf{F}^{-1}
%\frac{\partial f(\theta_t)}{\partial \theta_t}^T\right) -  f(\theta_t)} {
%-\frac{\partial f(\theta_t)}{\partial \theta_t} \mathbf{F}^{-1} \frac{\partial
%f(\theta_t)} {\partial \theta_t}^T - \frac{1}{2}\frac{\partial
%f(\theta_t)}{\partial \theta_t} \mathbf{F}^{-1} \frac{\partial f(\theta_t)}
%{\partial \theta_t}^T} \end{equation} which is just a constant away from eq.
%\ref{eq:our_eq}, implying that in principle using the same heuristic for
%damping the metric matrix is well motivated from a theoretical point of view.

\subsection*{Additional experimental results}

\subsubsection*{TFD experiment}

We used a two layer model, where the first layer is convolutional. It uses 512
filters of 14X14, and applies a sigmoid activation function. The next layer is
a dense sigmoidal one with  1024 hidden units. The output layer uses sigmoids
as well instead of softmax. 

The data is pre-processed by using local contrast normalization.
Hyper-parameters such as learning rate, starting damping coefficient have been
selected using a grid search, based on the validation cost obtained for each
configuration.

We ended up using a fixed learning rate of 0.2 (with no line search) and
adapting the damping coefficient using the Levenberg-Marquardt heuristic. 

When using the same samples for evaluating the metric and gradient we used
minibatches of 256 samples, otherwise we used 384 other samples randomly picked
from either the training set or the unlabeled set. We use MinResQLP as a linear
solver, the picked initial damping factor is 5., and we allow a maximum of 50
iterations to the linear solver.

\subsubsection*{NISTP experiment (robustness to the order of training samples)}

The model we experimented with was an MLP of only 500 hidden units. We compute
the gradients for both MSGD and natural gradient descent over minibatches of
512 examples. In the case of natural gradient descent we compute the metric
over the same input batch of 512 examples.  Additionally we use a constant
damping factor of 3 to account for the noise in the metric (and
ill-conditioning since we only use batches of 512 samples). The learning rates
were kept constant, and we use 0.2 for the natural gradient descent and 0.1 for
MSGD. 

\subsection*{Curves experiment - Benchmark}

For the curves experiment we used a deep autoencoder with 400-200-100-50-25-6
hidden units respectively where we apply sigmoid activation function everywhere
except on the top layer. We used the sparse initialization proposed by
\citet{Martens10} for all runs.

For natural conjugate gradient we use the full batch of 20000 examples to
compute the gradients or metric as well as for evaluating the model when
optimizing for $\alpha$ and $\beta$ using scipy.optimize.fmin\_cobyla.  We use
the Levenberg-Marquardt heuristic to adapt damping (and start with a damping
value of 3). We use MinRes-QLP that we allow to run for a maximum of 250 steps.
Batch size, number of linear solver iterations and damping coefficient have
been selected by hand tuning, looking at validation set performance.

For natural gradient we end up reporting two runs. One uses full batch and a
line search while the other uses small batches of 5000 examples and a fixed
learning rate of $1.0$. 

For the SGD case we use a smaller batch size of 100 examples.  The optimum
learning rate, obtained by a grid search, is 0.01 which was kept constant
through training. We did not use any additional enhancement like momentum.

A final note is that we show square error (to be consistent with
\citet{Martens10}) we minimize cross-entropy in practice (i.e.  the gradients
are computed based on the cross-entropy cost, square error is only computed for
visualization reasons).

\end{document}